\documentclass[11pt]{article}

\usepackage[utf8]{inputenc}
\usepackage[T1]{fontenc}
\usepackage{xcolor}
\usepackage{graphicx}
\usepackage{subfig}
\usepackage{wrapfig}
\usepackage{hyperref}
\usepackage{url}
\usepackage{booktabs} 
\usepackage{abstract}
\usepackage{multicol}
\usepackage{fancyhdr}
\usepackage{amsmath}
\usepackage{amsfonts}
\usepackage{amsthm}
\usepackage{amssymb}
\usepackage{nicefrac}
\usepackage{microtype}
\usepackage{paralist}
\usepackage[font=small,labelfont=bf]{caption}
\usepackage[ruled,norelsize]{algorithm2e}
\usepackage[noend]{algorithmic}
\usepackage{calc}
\usepackage{tikz}
\usepackage{bbm}
\usepackage{enumitem}
\usepackage{scalerel}
\usepackage{float}

\newcommand{\N}{\mathbb{N}}
\newcommand{\R}{\mathbb{R}}
\newcommand{\Normal}{\mathcal{N}}

\newcommand{\Ortho}{\mathrm{O}}

\newcommand{\bmat}{\begin{pmatrix}}
\newcommand{\emat}{\end{pmatrix}}

\title{The Hessian Estimation Evolution Strategy}

\fancypagestyle{firststyle}
{
   \fancyhf{}
   \fancyfoot{}
}

\author{
	Tobias Glasmachers\\
	Institute for Neural Computation, Ruhr-University Bochum, Germany\\
	\texttt{tobias.glasmachers@ini.rub.de}\\[0.5em]
	Oswin Krause\\
	Department of Computer Science, University of Copenhagen, Denmark\\
	\texttt{oswin.krause@di.ku.dk}
}

\date{}

\begin{document}

\maketitle

\setlength{\absleftindent}{2cm}
\setlength{\absrightindent}{2cm}

\begin{abstract}
\normalsize
We present a novel black box optimization algorithm called \emph{Hessian
Estimation Evolution Strategy}. The algorithm updates the covariance
matrix of its sampling distribution by directly estimating the curvature
of the objective function. This algorithm design is targeted at twice
continuously differentiable problems. 
For this, we extend the cumulative step-size adaptation algorithm of the CMA-ES to mirrored sampling.
We demonstrate that our approach to covariance matrix adaptation is
efficient by evaluating it on the BBOB/COCO testbed. We also show that
the algorithm is surprisingly robust when its core assumption of a twice
continuously differentiable objective function is violated.
The approach yields a new evolution strategy with competitive
performance, and at the same time it also offers an interesting
alternative to the usual covariance matrix update mechanism.
\end{abstract}

\section{Introduction}

We consider minimization of a black-box objective function
$f : \R^d \to \R$. Modern evolution strategies (ESs) are highly tuned
solvers for such problems \cite{hansen2010BBOB,hansen2016COCO}. The
state of the art is marked by the covariance matrix adaptation evolution
strategy (CMA-ES) \cite{hansen:2001,kern:2004} and its many variants.

Most modern evolution strategies (ESs) sample offspring from a Gaussian
distribution $\Normal(m, \sigma^2 C)$ around a single mean $m \in \R^d$.
Their most crucial mechanism is adaptation of the step size $\sigma >
0$, which enables them to converge at a linear rate on scale-invariant
problems \cite{jebalia2010log}. Hence they achieve the fastest possible
convergence speed class that can be realized by any comparison-based
algorithm \cite{teytaud2006general}. However, for ill-conditioned
problems the actual convergence rate can be very slow, i.e., the
multiplicative progress per step can be arbitrarily close to one. The
main role of CMA is to mitigate this problem: after successful
adaptation of the covariance matrix $C$ to a multiple of the inverse of
the Hessian $H$ of the problem, the ES makes progress at its optimal
rate.

To this end consider a convex quadratic function
$f(x) = \frac12 x^T H x$. Its Hessian matrix $H$ encodes the curvature
of the graph of $f$. Knowledge of this curvature is valuable for
optimization, e.g., turning a simple gradient step
$x \leftarrow x - \eta \cdot \nabla f(x)$ into a Newton step
$x \leftarrow x - H^{-1} \nabla f(x)$, which jumps straight into the
optimum. For an evolution strategy, adapting $C$ to $H^{-1}$ is
equivalent to learning a transformation of the input space that turns a
convex quadratic function into the sphere function. This way, after
successful adaptation, all convex quadratic functions are as easy to
minimize as the sphere function $f(x) = \frac12 \|x\|^2$, i.e., as if
the Hessian matrix was $H = I$ (the identity matrix). Due to Taylor's
theorem, the advantage naturally extends to local convergence into twice
continuously differentiable local optima, covering a large and highly
relevant class of problems.

The usual mechanism for adapting the covariance matrix $C$ of the
offspring generating distribution is to change it towards a weighted
maximum likelihood estimate of successful steps \cite{kern:2004}. The
update can equally well be understood as following a stochastic natural
gradient in parameter space \cite{ollivier2017IGO}.

In this paper we explore a conceptually different and more direct
approach for learning the inverse Hessian. It amounts to estimating the
curvature of the objective function on random lines through $m$ by means
of finite differences. We design a novel CMA mechanism for updating the
covariance matrix $C$ based on the estimated curvature information. We
call the resulting algorithm \emph{Hessian Estimation Evolution
Strategy} (HE-ES).

It is worth pointing out that estimating derivatives destroys an
important property of CMA-ES, namely invariance under strictly monotonic
transformations of objective values. Our new algorithm is still fully
invariant under order-preserving affine transformations of objective
values. This is an essentially equally good invariance guarantee only in
a local situation, namely if the value transformation is well approximated
by its first order Taylor polynomial. We address this potential weakness
in our experimental evaluation.

\noindent The main goals of this paper are
\begin{compactitem}
\item[$\bullet$]
	to present HE-ES and our novel CMA mechanism, and
\item[$\bullet$]
	to demonstrate its competitiveness with existing algorithms. To this
	end we compare HE-ES to CMA-ES as a natural baseline. We also
	compare with NEWUOA \cite{powell2004NEWUOA}, which is based directly
	on iterative estimates of a quadratic model of the objective
	function. Furthermore, we include BFGS \cite{nocedal2006numerical},
	a ``work horse'' (gradient-based) non-linear optimization algorithm.
	It is of interest because being a quasi-Newton method, it implicitly
	estimates the Hessian matrix.
\item[$\bullet$]
	Finally, we adapt cumulative step size adaptation to mirrored sampling.
\end{compactitem}
The remainder of the paper is structured as follows. In the next section
we describe the new algorithm in detail and briefly discuss its relation
to CMA-ES. Our main results are of empirical nature, therefore we
present a thorough experimental evaluation of the optimization
performance of HE-ES and discuss strengths and limitations. We close
with our conclusions.

\section{The Hessian Estimation Evolution Strategy}

The HE-ES algorithm is designed in as close as possible analogy to
CMA-ES. Ideally we would change only the covariance matrix adaptation
(CMA) mechanism. However, we end up changing also the offspring
generation method to a scheme that is tailored to estimating curvature
information. In the following we present the algorithm and motivate and
detail all mechanisms that deviate from CMA-ES~\cite{kern:2004}.

\paragraph{Estimating Curvature.~}

A seemingly natural strategy for estimating the Hessian of an unknown
black-box function is to estimate single entries $H_{ij}$ of this $d
\times d$ matrix by computing finite differences. For a diagonal entry
$H_{ii}$ this requires evaluating three points on a line parallel to the
$i$-th coordinate axis, while for an off-diagonal entry $H_{ij}$ we can
evaluate four corners of a rectangle with edges parallel to the $i$-th
and $j$-th coordinate axis. A serious problem with such a procedure is
that entries must remain consistent, which makes it difficult to design
an online update of a previous estimate of the matrix. Furthermore, the
scheme implies that offspring must be sampled along the coordinate axes,
which can significantly impair performance, e.g., on ill-conditioned
non-separable problems---which are exactly the problems we would like to
excel on.

We therefore propose a different solution. To this end we draw mirrored
samples $m \pm v$ and evaluate $f(m + \alpha v)$ for $\alpha \in \{-1,0,+1\}$.
Here $v \in \R^d$ is the direction of a line through $m$. The length
$\|v\|$ of that vector controls the scale on which the finite difference
estimate is computed. An estimate of the second directional derivative
of $f$ in direction $\frac{v}{\|v\|}$ at $m$ is
\begin{align*}
	\frac{f(m+v) + f(m-v) - 2 f(m)}{\|v\|^2}
	\enspace.
\end{align*}
For a convex quadratic function $f(x) = \frac12 (x-x^*)^T H (x-x^*)$ the
above quantity coincides with the directional derivative
$\frac{v^T H v}{\|v\|^2}$. It can be understood as the ``component'' of
$H$ in direction $\frac{v}{\|v\|}$. The expression simplifies to a
diagonal entry $H_{ii}$ if $v$ is a multiple of the $i$-th standard
basis vector $e_i = (0, \dots, 0, 1, 0, \dots, 0)$. For a general twice
continuously differentiable function the estimate converges to the above
value in the limit $\|v\| \rightarrow 0$. Importantly, $H$ is uniquely
determined by these components, so given enough directions $v$ there is
no need for a sampling procedure that corresponds to estimating
off-diagonal entries~$H_{ij}$.

\paragraph{Orthogonal Mirrored Sampling.~}

A single pair of mirrored samples provides information on the curvature
in a single direction $v$. We learn nothing about the curvature in the
$d-1$ dimensional space orthogonal to $v$. To make best use of the
available information we should therefore sample the next pair of
mirrored samples orthogonal to $v$. We apply the following sampling
procedure for random orthogonal directions, which was first proposed in
\cite{wang2019mirrored}. We draw $d$ Gaussian vectors and record their
lengths. The vectors are then orthogonalized with the Gram-Schmidt
procedure. Placing the vectors into a $d \times d$ matrix yields an
orthogonal matrix uniformly distributed in the orthogonal group
$\Ortho(d, \R)$. We then rescale the vectors to their original lengths.

\floatname{algorithm}{Procedure}

\begin{algorithm}
\caption{sampleOrthogonal}
\label{procedure:sampleOrthogonal}
\begin{algorithmic}[1]
\STATE{\textbf{input} dimension $d$}
\STATE{$z_1, \dots, z_d \sim \Normal(0, I)$}
\STATE{$n_1, \dots, n_d \leftarrow \|z_1\|, \dots, \|z_d\|$}
\STATE{apply the Gram-Schmidt procedure to $z_1, \dots, z_d$}
\STATE{return $y_i = n_i \cdot z_i, \quad i = 1, \dots, d$}
\end{algorithmic}
\end{algorithm}

The sampling procedure applied for each block is defined in
Procedure~\ref{procedure:sampleOrthogonal}. It is applicable to up to
$d$ pairs of mirrored samples. In general we aim to generate $\tilde \lambda$
pairs of mirrored samples, which amounts to $\lambda = 2 \tilde \lambda$
offspring in total.
We therefore split the pairs into $B = \lceil \tilde \lambda / d \rceil$
blocks and apply the above procedure $B$ times. The resulting vectors
are denoted as $b_{ij}$, where $i \in \{1, \dots, B\}$ is the block
index and $j \in \{1, \dots, d\}$ is the index within each block.

\paragraph{Covariance Matrix Update.~}

We aim for an update that modifies an existing covariance matrix in an
online fashion. A seemingly straightforward strategy is to adapt the
matrix so that after the update it matches the curvature in the
sampled directions. This approach is followed in
\cite{leventhal2011randomized}, and later in \cite{stich2016variable}.
However, such a strategy disregards the fact that all multiples of the
inverse Hessian are optimal covariance matrices. In fact, the update
would destroy a perfect covariance matrix simply because it differs
from the inverse Hessian by a large factor.

Therefore our goal is to adapt the covariance matrix to the closest
\emph{multiple} of the inverse Hessian $H^{-1}$. To this end we only
change curvature values relative to each other: if the measured
curvature in direction $v_1$ is $10$ times larger than in direction
$v_2$ then we aim to ensure that the updated matrix represents this
relation. Otherwise we modify the matrix as little as possible. In
particular, we keep its determinant (encoding the global scale)
constant: if an eigenvalue is increased, then another one is decreased
accordingly.

The easiest way to achieve the above goals is by means of a
multiplicative update
\cite{glasmachers2010xNES,krause2015xCMAES,beyer2017simplify}. We
decompose the covariance matrix%
\footnote{The decomposition is never computed explicitly in the
algorithm. Instead it directly updates the factor $A$.}
into the form $C = A^T A$. The mirrored samples take the form
$x_{ij}^\pm = m \pm \sigma \cdot A \, b_{ij}$, resulting in the
curvature estimates $h_{ij}$, which approximate
$b_{ij}^T A^T H A \, b_{ij}$. The update takes the form
$A' \leftarrow A G$ and hence $C' \leftarrow G C G$, where $G$ is a
symmetric positive definite matrix.

In the following we apply the above considerations on curvature
estimation to the function $\tilde f(x) = f\big(A (x -m)\big)$ using the
direction vectors $v = \sigma \cdot b_{ij}$. The actual goal of
optimization is to adapt $m$ towards $x^*$. Turning $\tilde f$ into the
sphere function greatly facilitates that process. It is achieved by
adapting $A$ towards (a multiple of) any Cholesky factor of $H^{-1}$,
or in other words, by making all eigenvalues of $A^T H A$ coincide.

In order to understand the update we first review a simplified example.
Consider only two vectors $b_1$ and $b_2$. For simplicity assume that
they fulfill $\sigma \|b_i\| = 1$, and assume that the curvature
estimates $h_{ii}$ are exact because the function is convex quadratic.
Then the ideal $G$ has an eigenvalue of $\sqrt[4]{h_{22} / h_{11}}$ for
eigenvector $b_1$, an eigenvalue of $\sqrt[4]{h_{11} / h_{22}}$ for
eigenvector $b_2$, and eigenvalue $1$ in the space orthogonal to $b_1$
and $b_2$.
This seemingly very specific choice ensures that $\det(G)=1$ and it holds
\begin{align*}
	b_1^T (A')^T H A' \, b_1
	&= b_1^T G A^T H A G b_1 \\
	&= \sqrt{\frac{h_{22}}{h_{11}}} \cdot b_1^T A^T H A \, b_1
	= \sqrt{\frac{h_{22}}{h_{11}}} \cdot h_{11}
	= \sqrt{h_{11} h_{22}}
	\enspace,
\end{align*}
which coincides with $b_2^T (A')^T H A' b_2$ for symmetry reasons.
Hence, after the update the curvatures in directions $b_1$ and $b_2$
have become equal, while all curvatures orthogonal to the sampling
directions remain unchanged. In this sense the resulting problem
$\tilde f$ has come closer to the sphere function.

\begin{algorithm}
\caption{computeG}
\label{procedure:computeG}
\begin{algorithmic}[1]
\STATE{\textbf{input} $b_{ij}$, $f(m)$, $f(x_{ij}^\pm)$, $\sigma$}
\STATE{\textbf{parameters} $\kappa$, $\eta_A$}
\STATE{$h_{ij} \leftarrow \frac{f(x_{ij}^+) + f(x_{ij}^-) - 2 f(m)}{\sigma^2 \cdot \|b_{ij}\|^2}$ \hfill \# estimate curvature along $b_{ij}$}
\STATE{\textbf{if} $\max(\{h_{ij}\}) \leq 0$ \textbf{then} \textbf{return} $I$}
\STATE{$c \leftarrow \max(\{h_{ij}\}) / \kappa$}
\STATE{$h_{ij} \leftarrow \max(h_{ij}, c)$ \hfill \# truncate to trust region}
\STATE{$q_{ij} \leftarrow \log(h_{ij})$}
\STATE{$q_{ij} \leftarrow q_{ij} - \frac{1}{\tilde \lambda} \cdot \sum_{ij} q_{ij}$ \hfill \# subtract mean $\to$ ensure unit determinant}
\STATE{$q_{ij} \leftarrow q_{ij} \cdot \frac{-\eta_A}{2}$ \hfill \# learning rate and inverse square root (exponent $-1/2$)}
\STATE{$q_{B,j} \leftarrow 0 \quad \forall j \in \{d B - \tilde \lambda, \dots, d\}$ \hfill \# neutral update in the unused directions}
\STATE{\textbf{return} $\frac{1}{B} \sum_{ij} \frac{\exp(q_{ij})}{\|b_{ij}\|^2} \cdot b_{ij} b_{ij}^T$}
\end{algorithmic}
\end{algorithm}

A generalization of the above update to an arbitrary number of
(unnormalized) update directions $\sigma \cdot b_{ij}$ is implemented
by Procedure~\ref{procedure:computeG}, which computes the matrix $G$. It
forms the algorithmic core of our method.

This core works very well, for example, for smooth convex problems. General
non-smooth and non-convex objective functions can exhibit unstable curvature
estimates $h_{ij}$. A noisy objective function can create similar issues.
For non-convex problems, the eigenvalues of a Hessian can be zero or even
negative, in contrast to the eigenvalues of a covariance matrix. In all of
these situations we are better off to smoothen the update. Therefore
procedure~\ref{procedure:computeG} takes two measures: first,
it bounds the conditioning of the multiplicative update $G$ by a constant $\kappa$
to limit the effect of
outliers (lines 5-6), and second it applies a learning rate $\eta_A \in (0, 1]$
to stabilize estimates through temporal smoothing (line~9). The first of
these mechanisms also gracefully handles curvature estimates $h_{ij} \leq 0$
by clipping them to a small positive value. This is a reasonable thing to do
since we want to emphasize sampling in these directions without completely
destroying the learned information. In our experiments, we use the settings
$\kappa = 3$ and $\eta_A = 1/2$. They represent a reasonable compromise
between stability and adaptation speed.

\paragraph{Cumulative Step-size Adaptation (CSA) for Orthogonal Frames.~}
The usage of mirrored sampling for CSA was previously explored in
\cite{brockhoff2010mirrored,wang2019mirrored}. It was found that the default
algorithm exhibits a strong step size decay on flat or random function surfaces.
In the past this issue was alleviated by  not considering the mirrored samples
from the population when computing the CSA update. This however is inefficient
as only half of the
samples are used to update the step-size. In this section, we will
quantify the step-length bias of CSA with mirrored samples under selection
among all offspring. For this, we observe that under mirrored sampling each
direction $b_{ij}$ obtains two weights $w_{ij}^+$ and
$w_{ij}^-$ based on the function-values of the mirrored pair $x_{ij}^\pm$. Thus,
we can write the CSA mean computation \cite{hansen:2001} as
\begin{equation}\label{eq:CSAMean}
	\sum_{i,j}\left( w_{ij}^+ \frac {A^{-1}(x^+_{ij} - m)}{\sigma}+w_{ij}^- \frac {A^{-1}(x^-_{ij} - m)}{\sigma}\right)
	= \sum_{i,j} (w_{ij}^+ - w_{ij}^-) b_{ij}\enspace,
\end{equation}
leading to an update of the evolution path
\begin{equation}\label{eq:path}
    p_s^{(t+1)} \leftarrow (1 - c_s) \cdot p_s^{(t)} + \sqrt{c_s \cdot (2 - c_s) \cdot \mu_\text{eff}} \cdot \sum_{ij} (w_{ij}^+ - w_{ij}^-) \cdot b_{ij}\enspace.
\end{equation}
In the CSA, the evolution path is updated such that
its expected length under random selection is the expected value of the
$\chi^2(d)$ distribution. To correct for the bias introduced by the weighted mean, the CSA adds the correction $\mu_\text{eff} =1/\sum_i w^2_i$. In the mirror-sampling case, the subtraction of the weights
on the right hand side of \eqref{eq:CSAMean} means that the expected length of the vector is
smaller than expected under non-mirrored sampling, therefore the step-size update is biased and
tends to reduce the step-size prematurely. We fix this problem by
computing the correct normalization factor $\mu_\text{eff}^\text{mirrored} > \mu_\text{eff}$.

Under random selection, $w_{ij}^+$ and $w_{ij}^-$ are randomly picked without
replacement from $\vec w$, independently of $b_{ij}$. Thus, the distribution of the weighted sample-average of \eqref{eq:CSAMean} is still normal and the expected squared length is:
 
\begin{align*}
&E\left\{\left\lVert \sum_{i,j} (w_{ij}^+ - w_{ij}^-) b_{ij} \right\rVert^2 \right\} 
= E\left\{\sum_{i,j} (w_{ij}^+ - w_{ij}^-)^2 \right\} E\left\{\lVert y \rVert^2\right\} \\
&= E\left\{\sum_{i,j} (w_{ij}^+ - w_{ij}^-)^2\right\}d
\end{align*}
Note that in the first step, we used that $E\left\{y_i^Ty_j\right\}= 0$
for $i \neq j$, while the second step holds because we ensure during
sampling that the squared length of the samples is still
$\chi^2(d)$-distributed. Next, we will use that the set of all $w_{ij}^+$ and $w_{ij}^-$ together forms the weight vector $\vec w$ and thus the expectation can be written as permutations $\tau$ of the indices of $\vec w$. We can therefore write the expectation in terms of $w_i$ as:

\begin{align*}
    E\left\{\sum_{i,j}^{\tilde \lambda} (w_{ij}^+ - w_{ij}^-)^2\right\}
    &= E\left\{\sum_{i,j}^{\tilde \lambda} (w_{ij}^+)^2 + (w_{ij}^-)^2 - 2w_{ij}^+ w_{ij}^- \right\}\\
    &= E_{\tau}\left\{\sum_{k=1}^{\tilde \lambda} w_{\tau(k)}^2 + w_{\tau(k+\tilde \lambda)}^2  - 2w_{\tau(k)} w_{\tau(k+\tilde \lambda)} \right\}\\
    &= \frac 1 {\mu_{\text{eff}}}- 2E_{\tau}\left\{\sum_{k=1}^{\tilde \lambda} w_{\tau(k)} w_{\tau(k+\tilde \lambda)} \right\}
\end{align*}
To continue, we expand the expectation and count the number of times each $(w_i,w_j)$-pair appears. There is a total of $(2 \tilde \lambda)!$ permutations and for each $(i,j)$-pair with $i \neq j$, there are $ (2 \tilde \lambda)!/ (2 \tilde \lambda \cdot (2 \tilde \lambda-1))$ permutations such that $\tau(k)=i$ and $\tau(k+\tilde \lambda)=j$ for each $k$, which leads to another factor of $\tilde \lambda$. Thus, we obtain:
\begin{multline*}
     \frac 1 {\mu_{\text{eff}}}- 2E_{\tau}\left\{\sum_{k=1}^{\tilde \lambda} w_{\tau(k)}w_{\tau(k+\tilde \lambda)} \right\}
    = \frac 1 {\mu_{\text{eff}}}- \frac 2 {(2 \tilde \lambda)!}\sum_{i,j \neq i}^{2 \tilde \lambda} \frac{\tilde \lambda (2 \tilde \lambda)!}{2 \tilde \lambda \cdot (2 \tilde \lambda-1)}w_iw_j\\ 
    = \frac 1 {\mu_{\text{eff}}}- \frac 1 {2 \tilde \lambda-1}\sum_{i,j \neq i}^{2 \tilde \lambda}w_iw_j
    = \frac 1 {\mu_{\text{eff}}}- \frac 1 {2 \tilde \lambda-1}\sum_{i}^{2 \tilde \lambda}w_i(1-w_i)\\
    = \frac 1 {\mu_{\text{eff}}}- \frac 1 {2 \tilde \lambda-1}\left(1- \frac 1 {\mu_{\text{eff}}}\right)
    = \frac 1 {\mu_{\text{eff}}}\left(1 - \frac {\mu_{\text{eff}} - 1} {2 \tilde \lambda-1}\right)
\end{multline*}
In the second step, we use $\sum_{j \neq i}^{2 \tilde \lambda}w_j= 1-w_i$. Thus, we remove the step-length bias in CSA by replacing $\mu_{\text{eff}}$ in equation~\eqref{eq:path} with
\begin{align*}
	\mu_\text{eff}^\text{mirrored} := \frac{\mu_\text{eff}} {1 - \frac {\mu_\text{eff}  - 1}{2 \tilde \lambda - 1}}.
\end{align*}

\paragraph{The Algorithm.~}

\floatname{algorithm}{Algorithm}

\begin{algorithm}
\caption{Hessian Estimation Evolution Strategy (HE-ES)}
\label{algorithm:HE-ES}
\begin{algorithmic}[1]
\STATE{\textbf{input} $m^{(0)} \in \R^d$, $\sigma^{(0)} > 0$, $A^{(0)} \in \R^{d \times d}$}
\STATE{\textbf{parameters} $\tilde \lambda \in \N$, $c_s$, $d_s$, $w \in \mathbb{R}^{2 \tilde \lambda }$}
\STATE{$B \leftarrow \lceil \tilde \lambda / d \rceil$}
\STATE{$p_s^{(0)} \leftarrow 0 \in \R^d$}
\STATE{$g_s^{(0)} \leftarrow 0$}
\STATE{$t \leftarrow 0$}
\REPEAT
	\FOR{$j \in \{1, \dots, B\}$}
		\STATE{$b_{1j}, \dots, b_{dj} \leftarrow$ \texttt{sampleOrthogonal}()}
	\ENDFOR
	\STATE{$x_{ij}^- \leftarrow m^{(t)} - \sigma^{(t)} \cdot A^{(t)} b_{ij}$ ~~~~~for $i+(j-1)B \leq \tilde \lambda$}
	\STATE{$x_{ij}^+ \leftarrow m^{(t)} + \sigma^{(t)} \cdot A^{(t)} b_{ij}$ ~~~~~for $i+(j-1)B \leq \tilde \lambda$ \hfill \# mirrored sampling}
	\STATE{$A^{(t+1)} \leftarrow A^{(t)} \cdot$ \texttt{computeG}($\{b_{ij}\}$, $f(m)$, $\{f(x_{ij}^\pm)\}$, $\sigma^{(t)}$)} \hfill \# matrix adaptation
    \STATE{$w_{ij}^\pm  \leftarrow w_{\text{rank}(f(x_{ij}^\pm))}$}
	\STATE{$m^{(t+1)} \leftarrow \sum_{ij} w_{ij}^\pm \cdot x_{ij}^\pm$} \hfill \# mean update
	\STATE{$g_s^{(t+1)} \leftarrow (1 - c_s)^2 \cdot g_s^{(t)} + c_s \cdot (2 - c_s)$} \hfill
	\STATE{$p_s^{(t+1)} \leftarrow (1 - c_s) \cdot p_s^{(t)} + \sqrt{c_s \cdot (2 - c_s) \cdot \mu_\text{eff}^\text{mirrored}} \cdot \sum_{ij} (w_{ij}^+ - w_{ij}^-) \cdot b_{ij}$}
	\STATE{$\sigma^{(t+1)} \leftarrow \sigma^{(t)} \cdot \exp\left( \frac{c_s}{d_s} \cdot \frac{\|p_s^{(t+1)}\|}{\chi_d} - \sqrt{g_s^{(t+1)}} \right)$} \hfill \# CSA
	\STATE{$t \leftarrow t + 1$}
\UNTIL{ \textit{stopping criterion is met} }
\end{algorithmic}
\end{algorithm}

The resulting HE-ES algorithm is summarized in
Algorithm~\ref{algorithm:HE-ES}. Up to the (significant) changes
discussed in the previous sections its design is identical to CMA-ES. In
particular, it relies on global intermediate recombination, non-elitist
selection, cumulative step-size adaptation (CSA), and it applies the
same weights as CMA-ES to the offspring~\cite{hansen:2001}.
As default number of mirrored directions, we chose  $\tilde \lambda = 2 + \lfloor \frac 3 2 \log(d)\rfloor$.
As learning-rates $c_s$ and $d_s$ of the CSA in HE-ES, we chose the same
values as the implementation of the  CMA in \texttt{pycma-2.7.0} with
$2 \tilde \lambda$ offspring. In contrast to CMA-ES, HE-ES needs to
evaluate $f(m)$ in each generation. This value is used only for
estimating curvatures.

\section{Experimental Evaluation}

Our experimental evaluation aims to answer the following research
questions:
\begin{compactenum}
\item
	Is Hessian estimation a competitive CMA scheme?
\item
	What are its strengths and weaknesses compared with CMA-ES?
\item
	How much does performance change under monotonically increasing but
	non-affine fitness transformations?
\end{compactenum}
The source code of the algorithm that was used in all experiments is
available from the first author's website.%
\footnote{\texttt{https://www.ini.rub.de/the\_institute/people/tobias-glasmachers/\#software}}

\paragraph{Benchmark Study.~}

Our first experiment is to run the standardized BBOB/COCO procedure,
which tests the algorithm on 15 instances of 24 benchmark problems
\cite{hansen2016COCO}. For handling multi-modal problems we equip HE-ES
with an IPOP restart mechanism \cite{auger2005restart}, which restarts
the algorithm with doubled population size as soon the standard
deviation of the fitness values of a generation falls below $10^{-9}$.

\begin{figure*}
\includegraphics[width=0.32\textwidth]{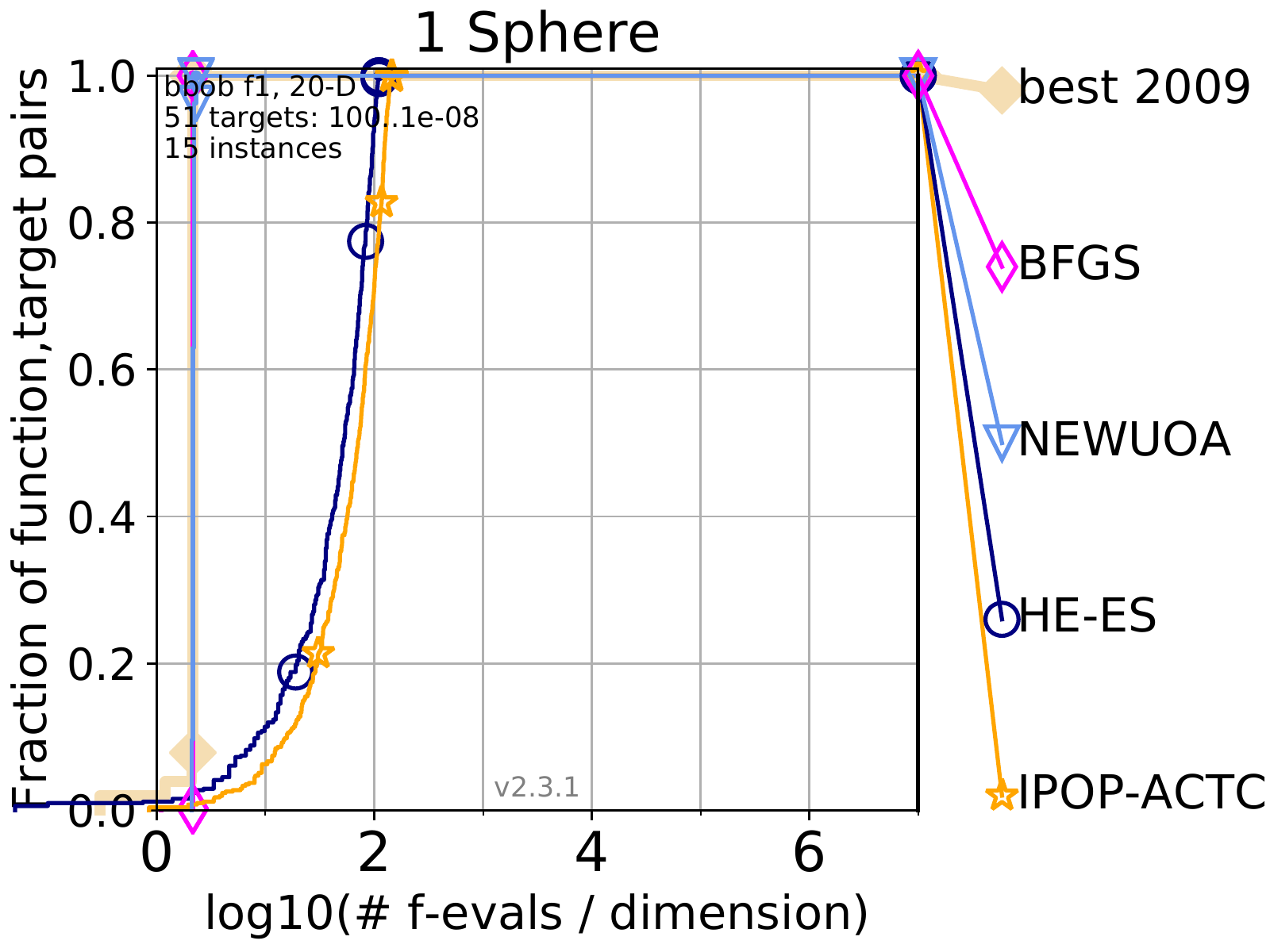}%
\includegraphics[width=0.32\textwidth]{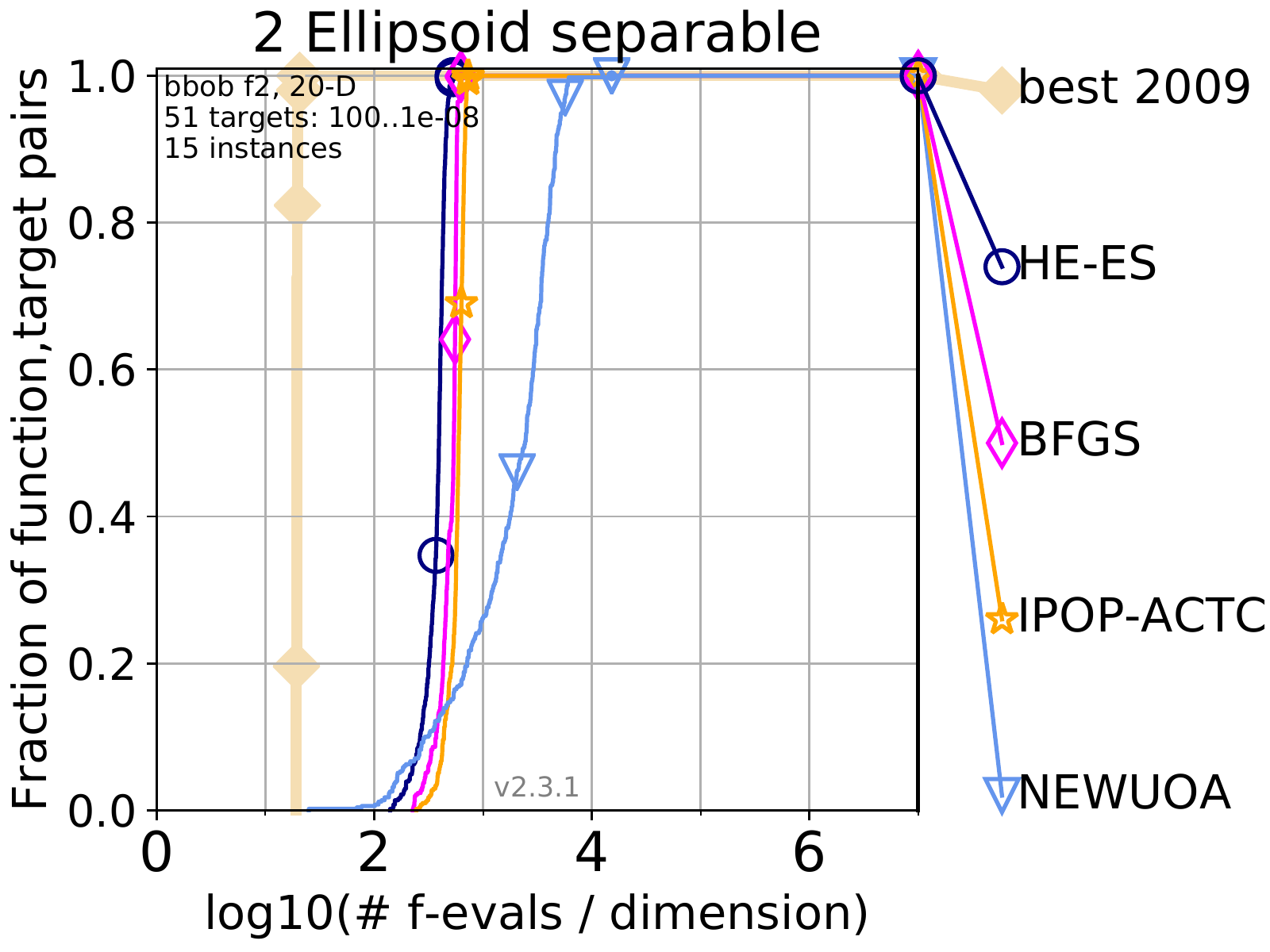}%
\includegraphics[width=0.32\textwidth]{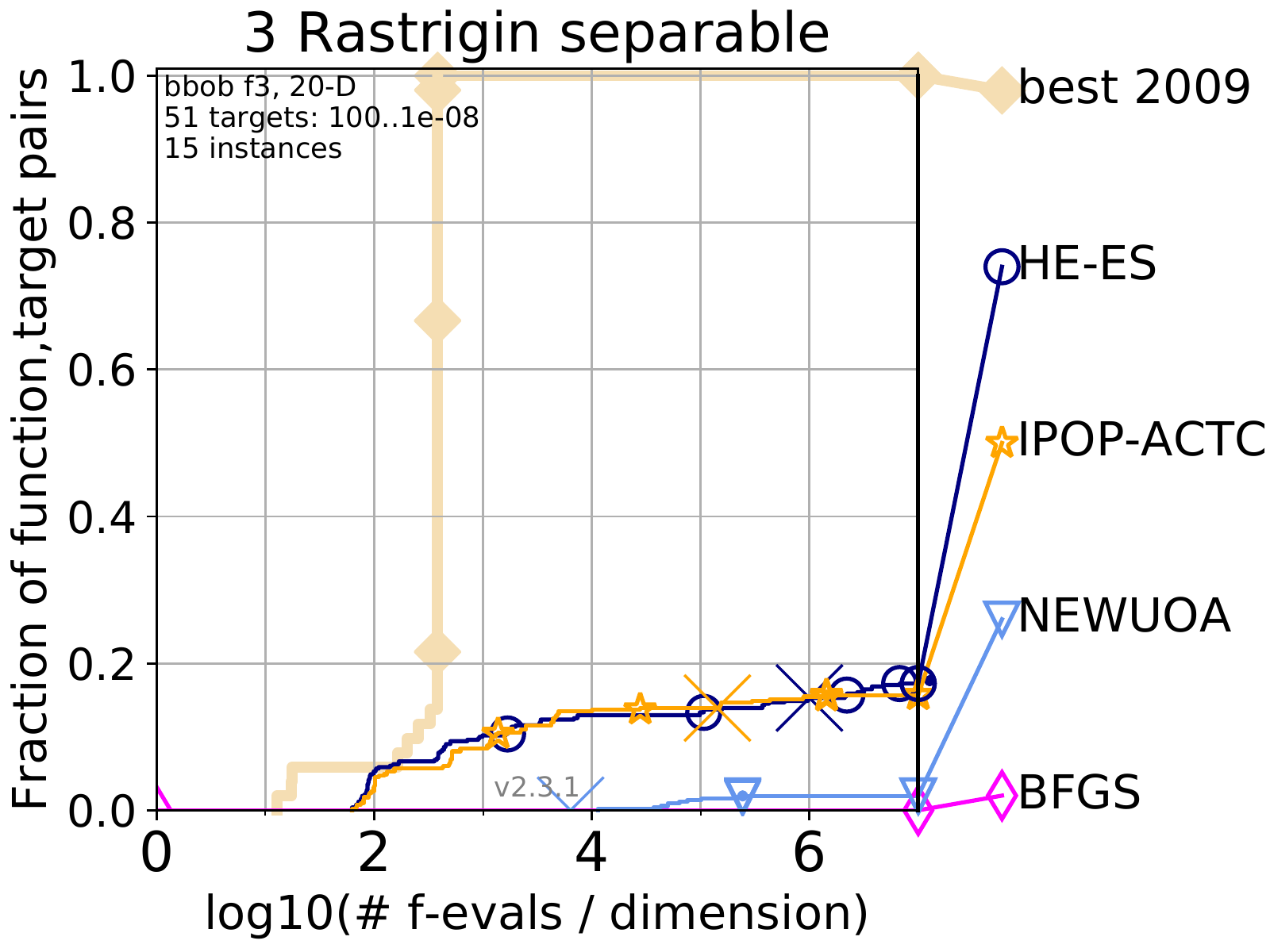}%
\\
\includegraphics[width=0.32\textwidth]{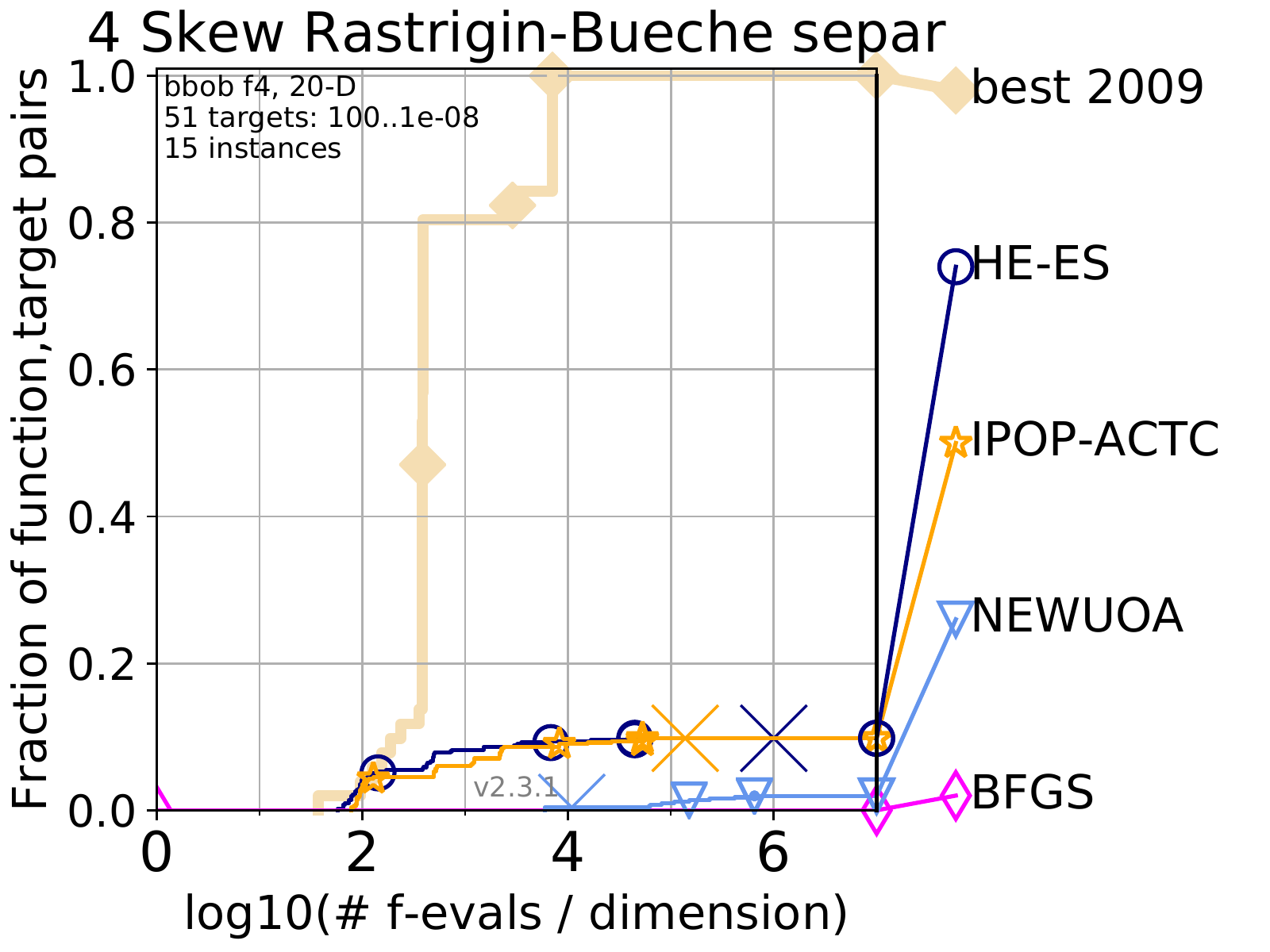}%
\includegraphics[width=0.32\textwidth]{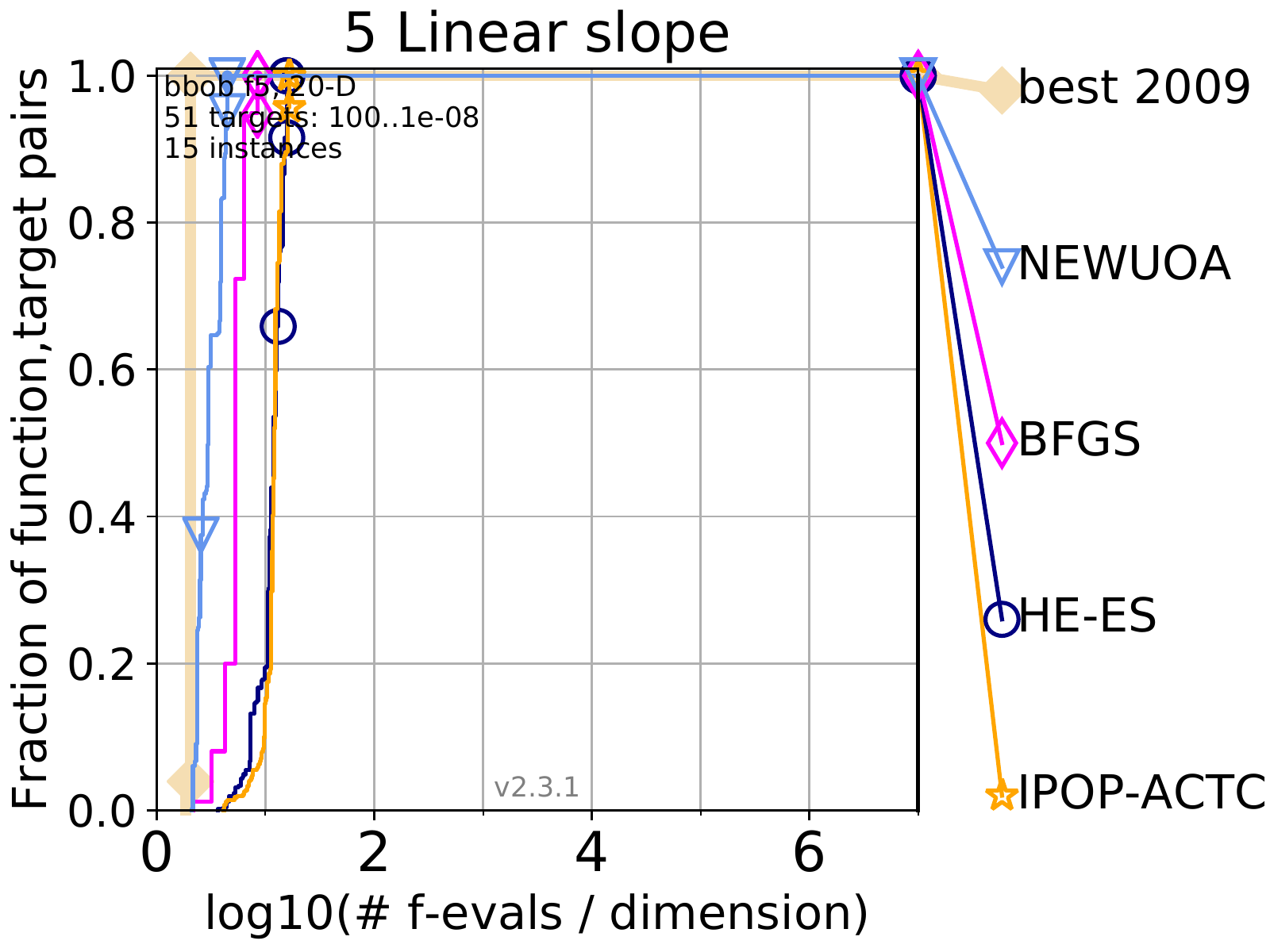}%
\includegraphics[width=0.32\textwidth]{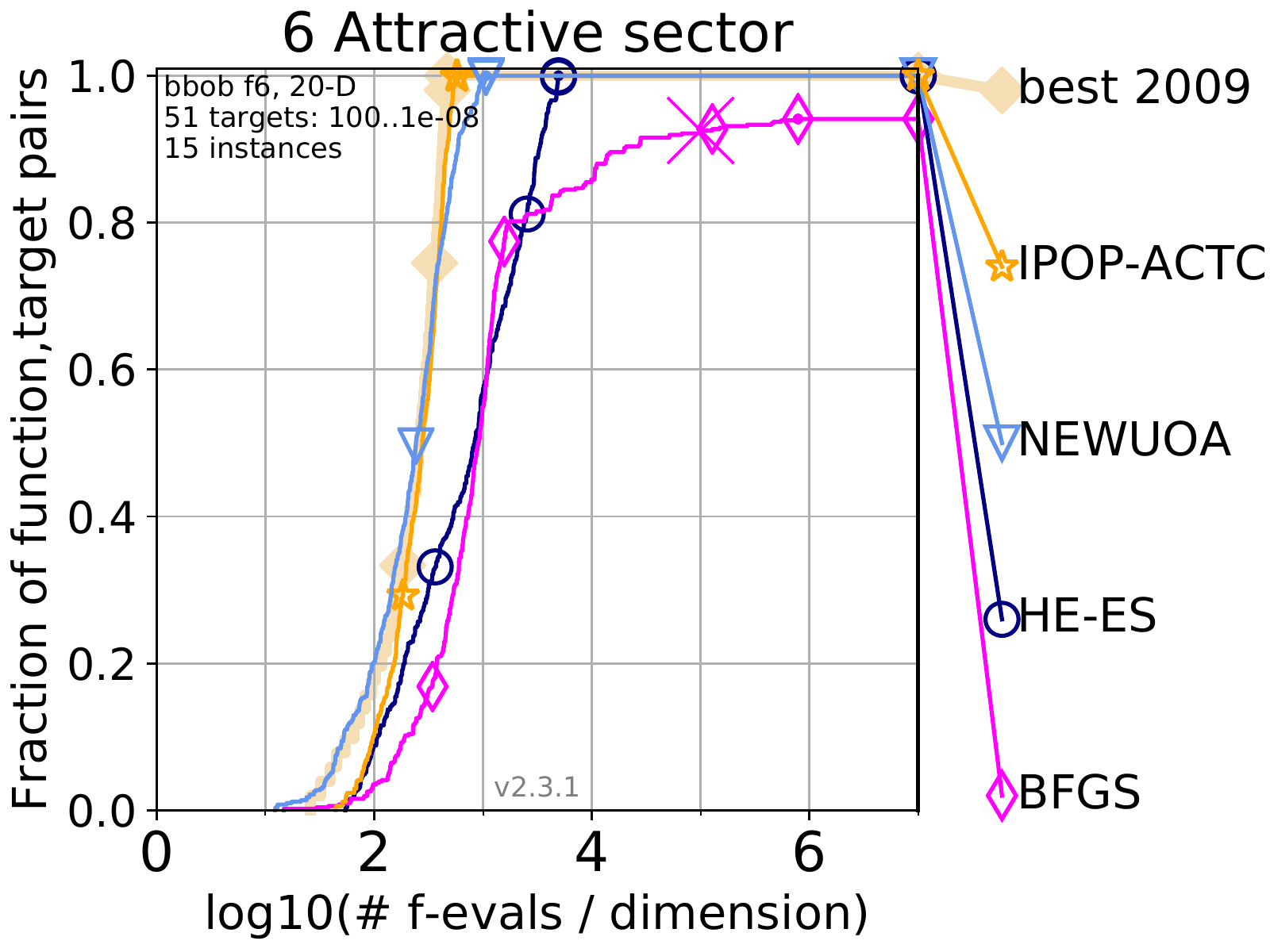}%
\\
\includegraphics[width=0.32\textwidth]{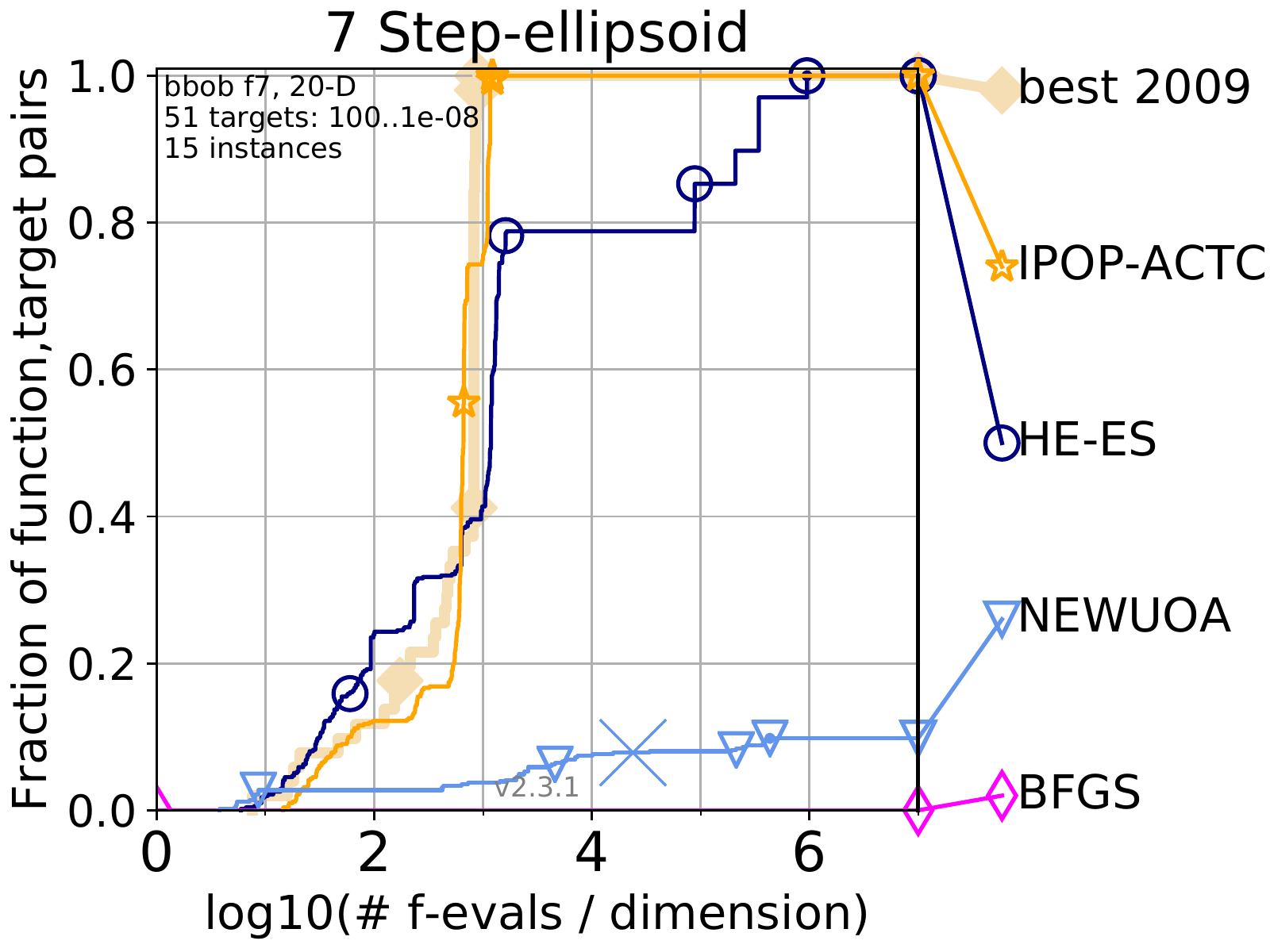}%
\includegraphics[width=0.32\textwidth]{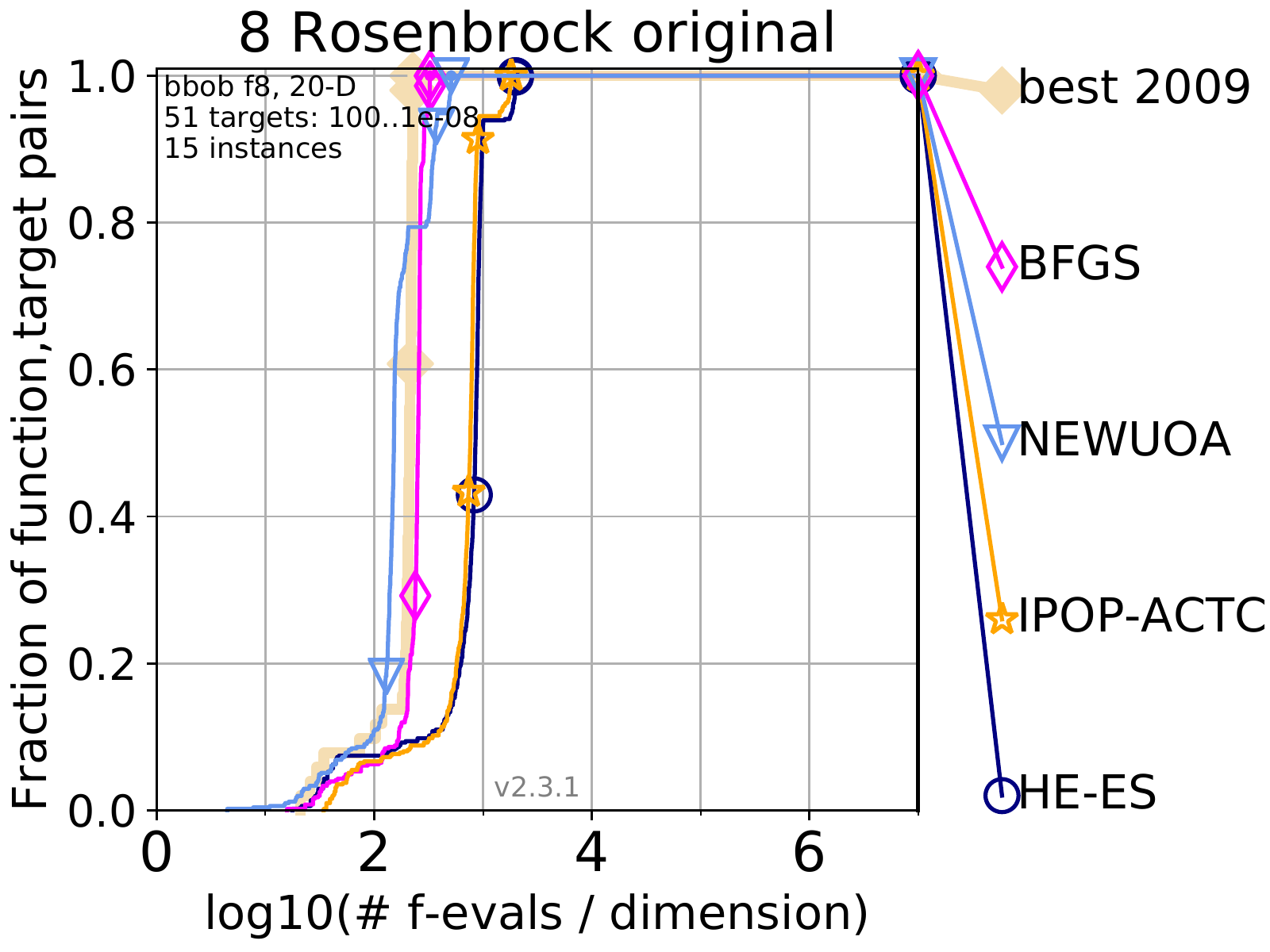}%
\includegraphics[width=0.32\textwidth]{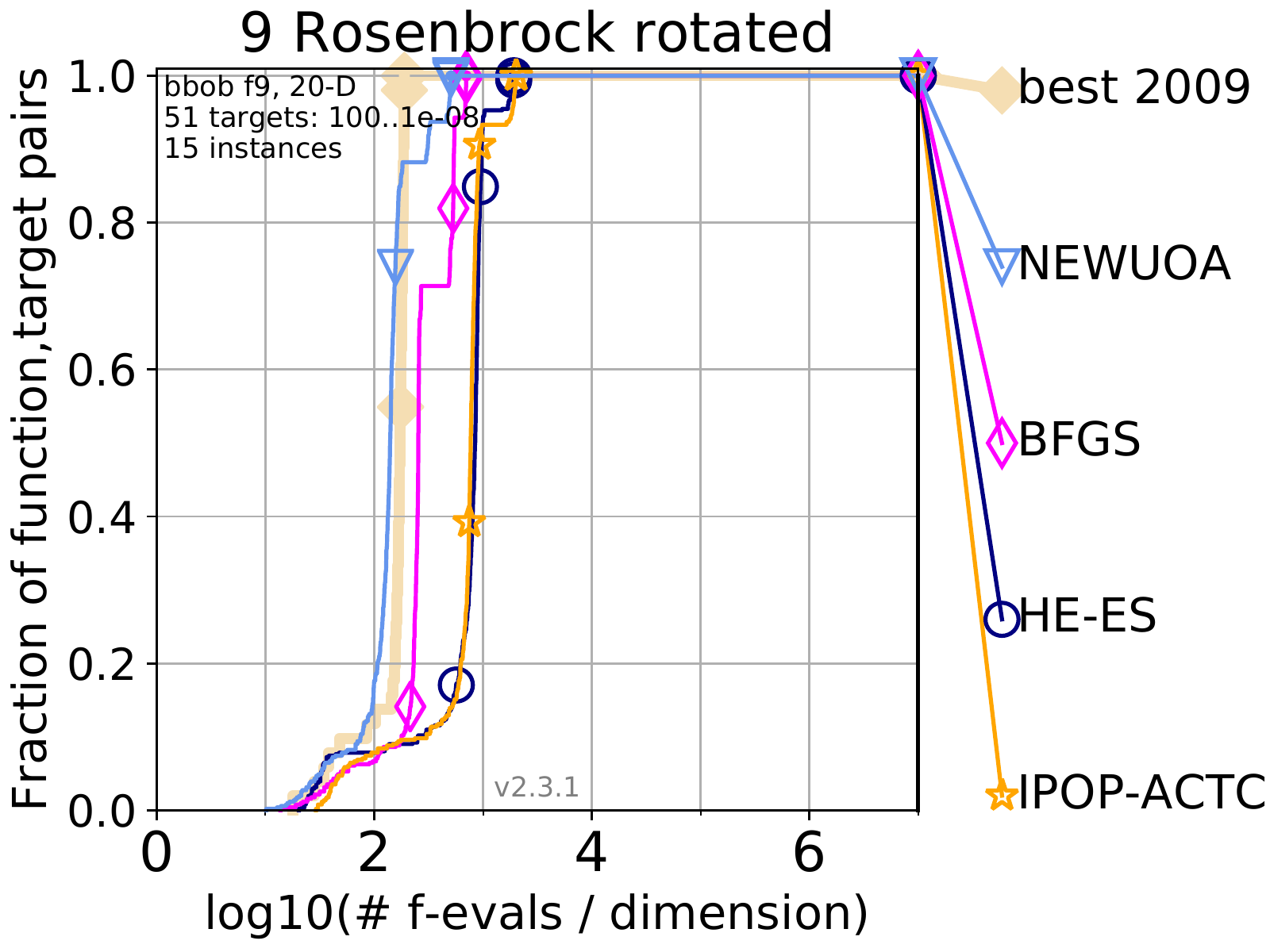}%
\\
\includegraphics[width=0.32\textwidth]{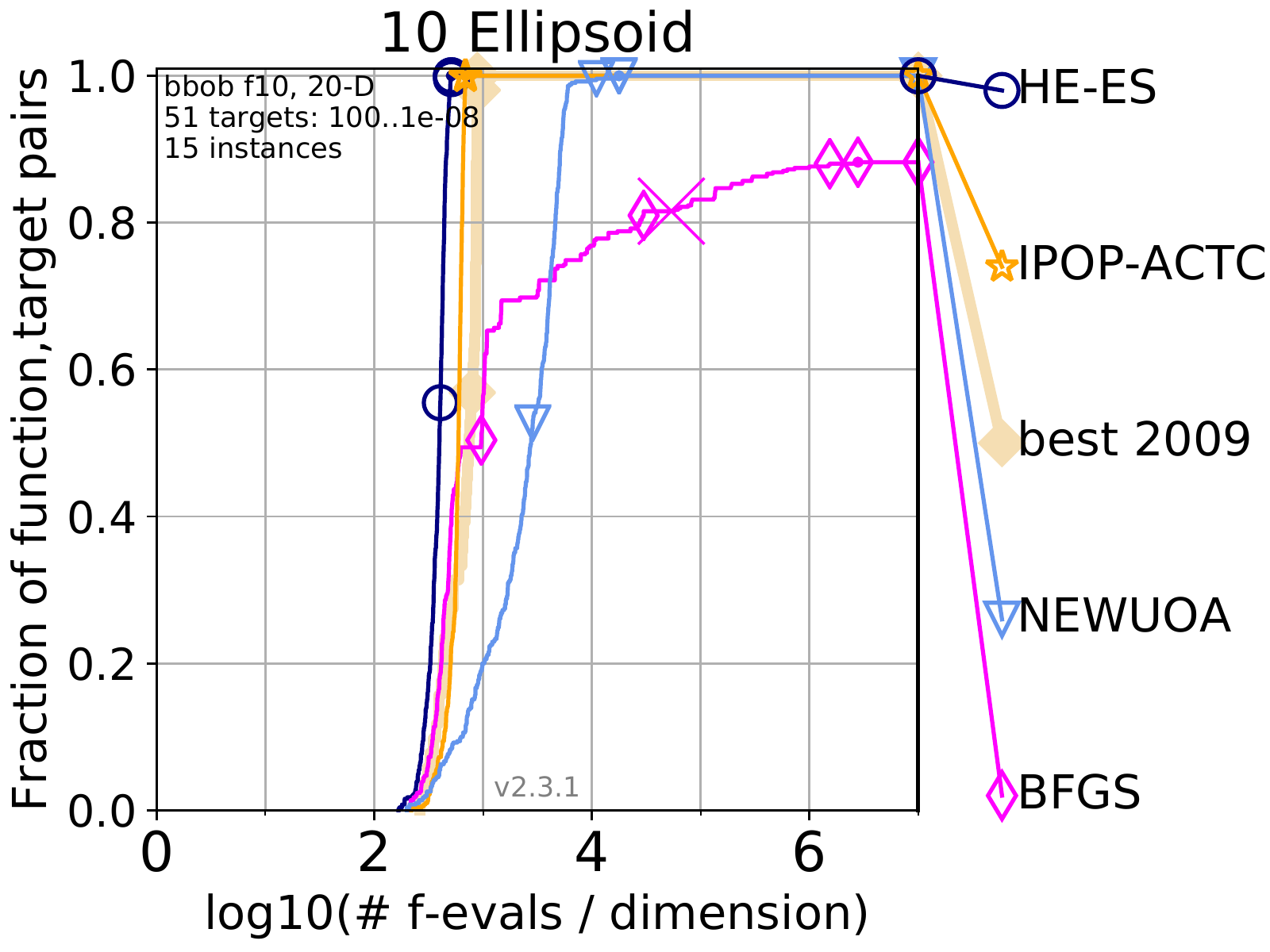}%
\includegraphics[width=0.32\textwidth]{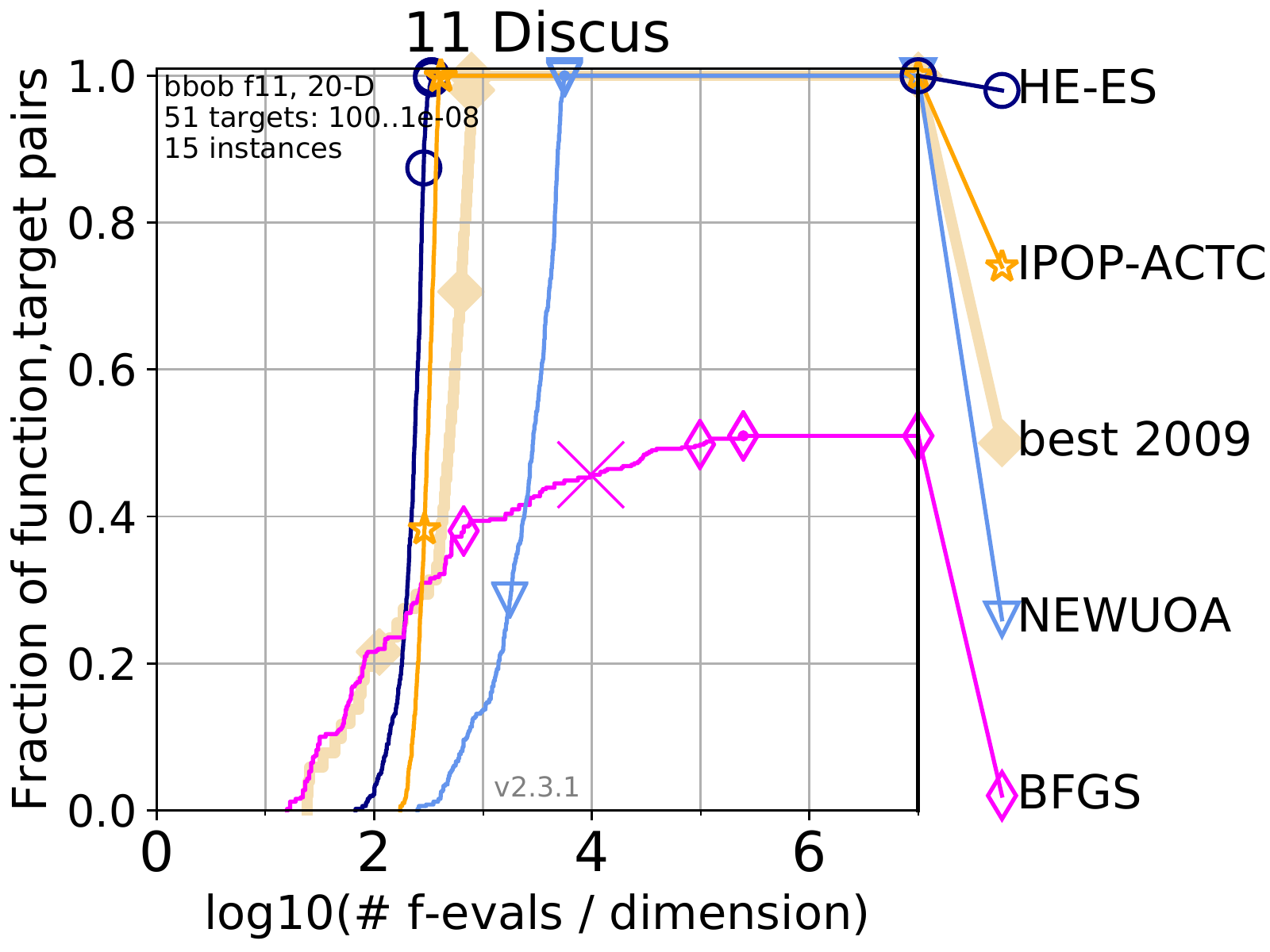}%
\includegraphics[width=0.32\textwidth]{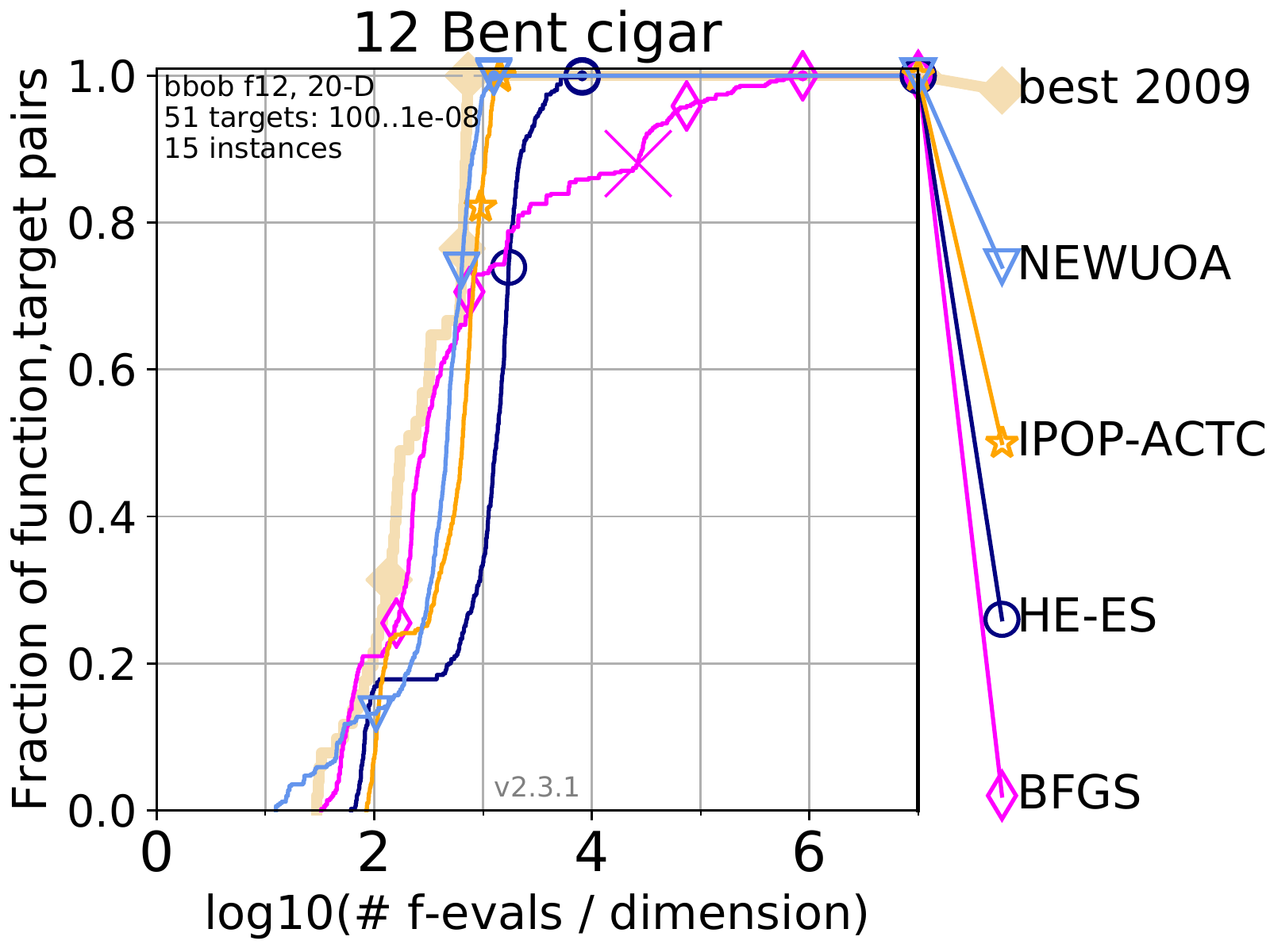}%
\\
\includegraphics[width=0.32\textwidth]{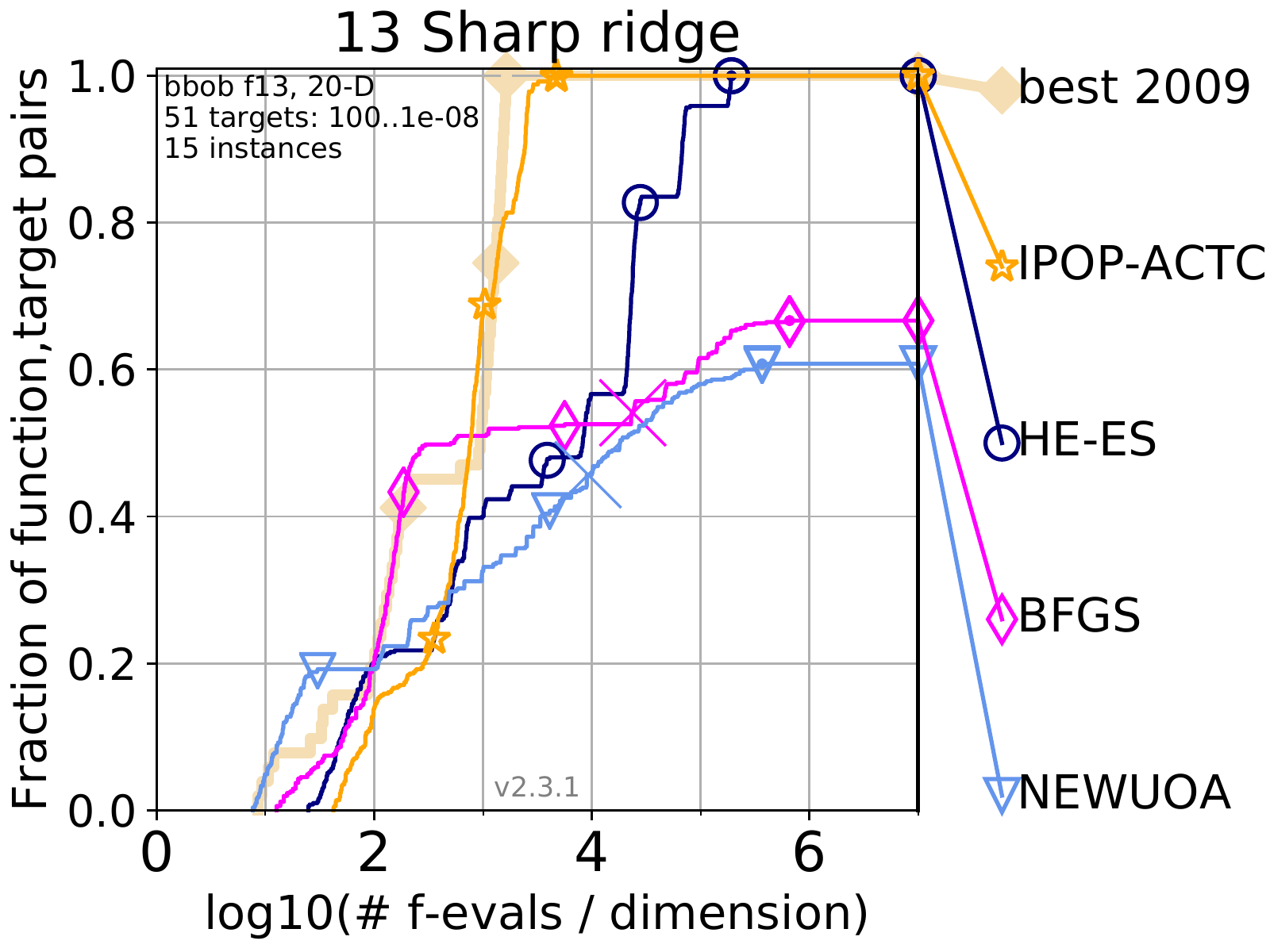}%
\includegraphics[width=0.32\textwidth]{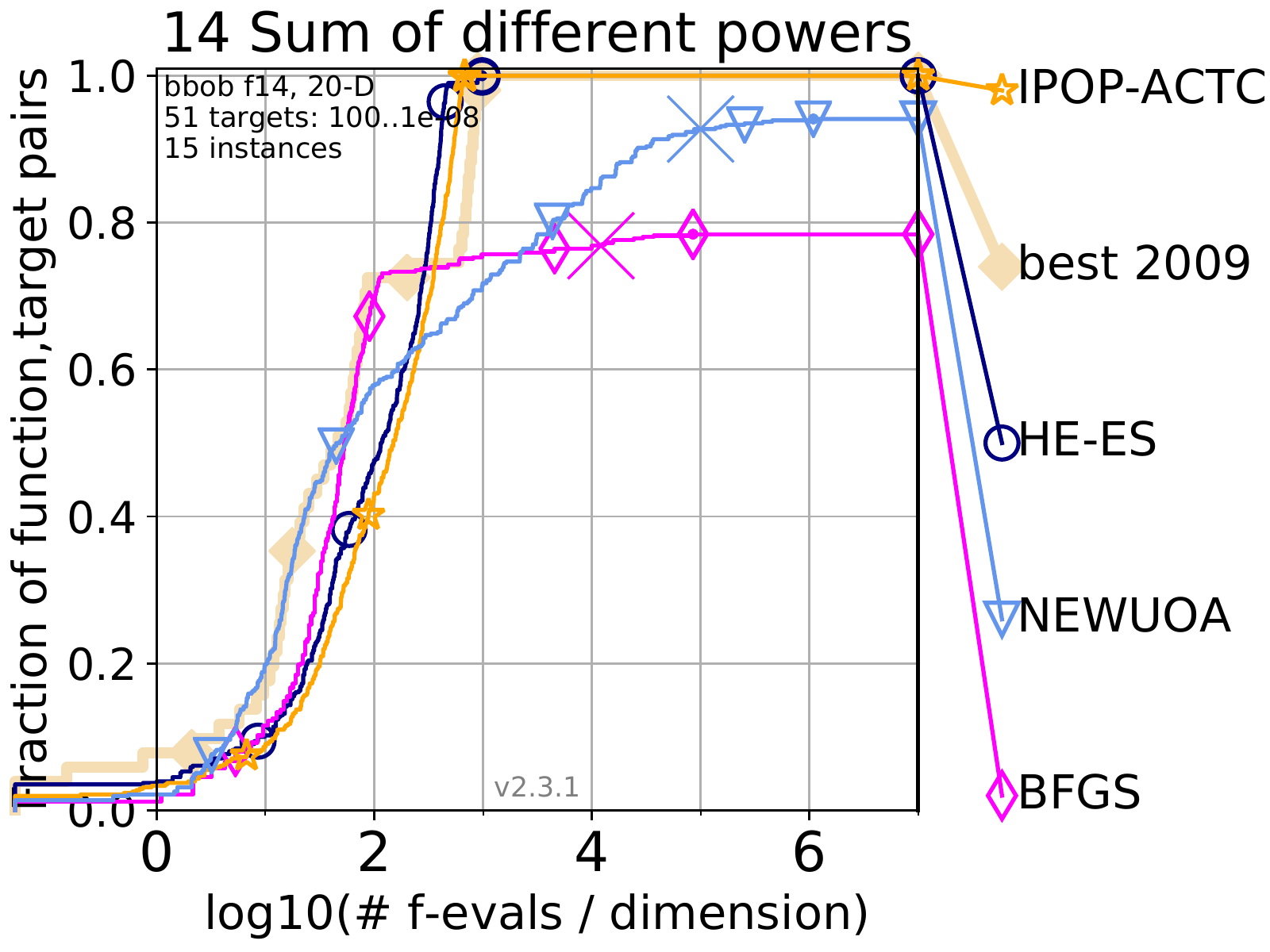}%
\includegraphics[width=0.32\textwidth]{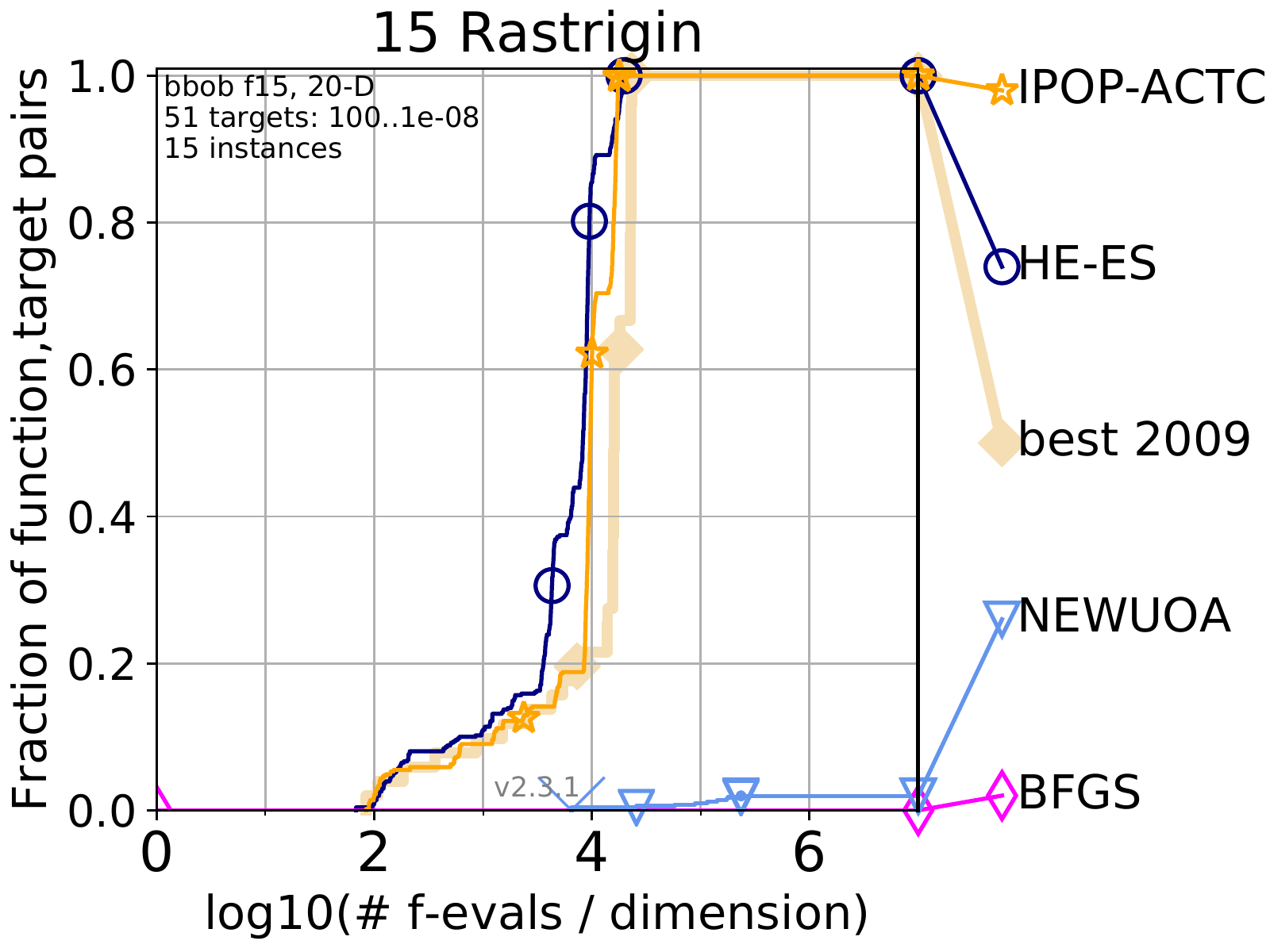}%
\caption{
	ECDF plots for the noiseless BBOB problems 1--15 in dimension
	$d=20$. We generally observe that HE-ES performs very well on
	smooth unimodal problems (functions 1, 2, 5, 8, 9, 10, 11, 12, 14).
\label{figure:ecdf}}
\vspace*{-1pt} 
\end{figure*}

\begin{figure*}
\includegraphics[width=0.32\textwidth]{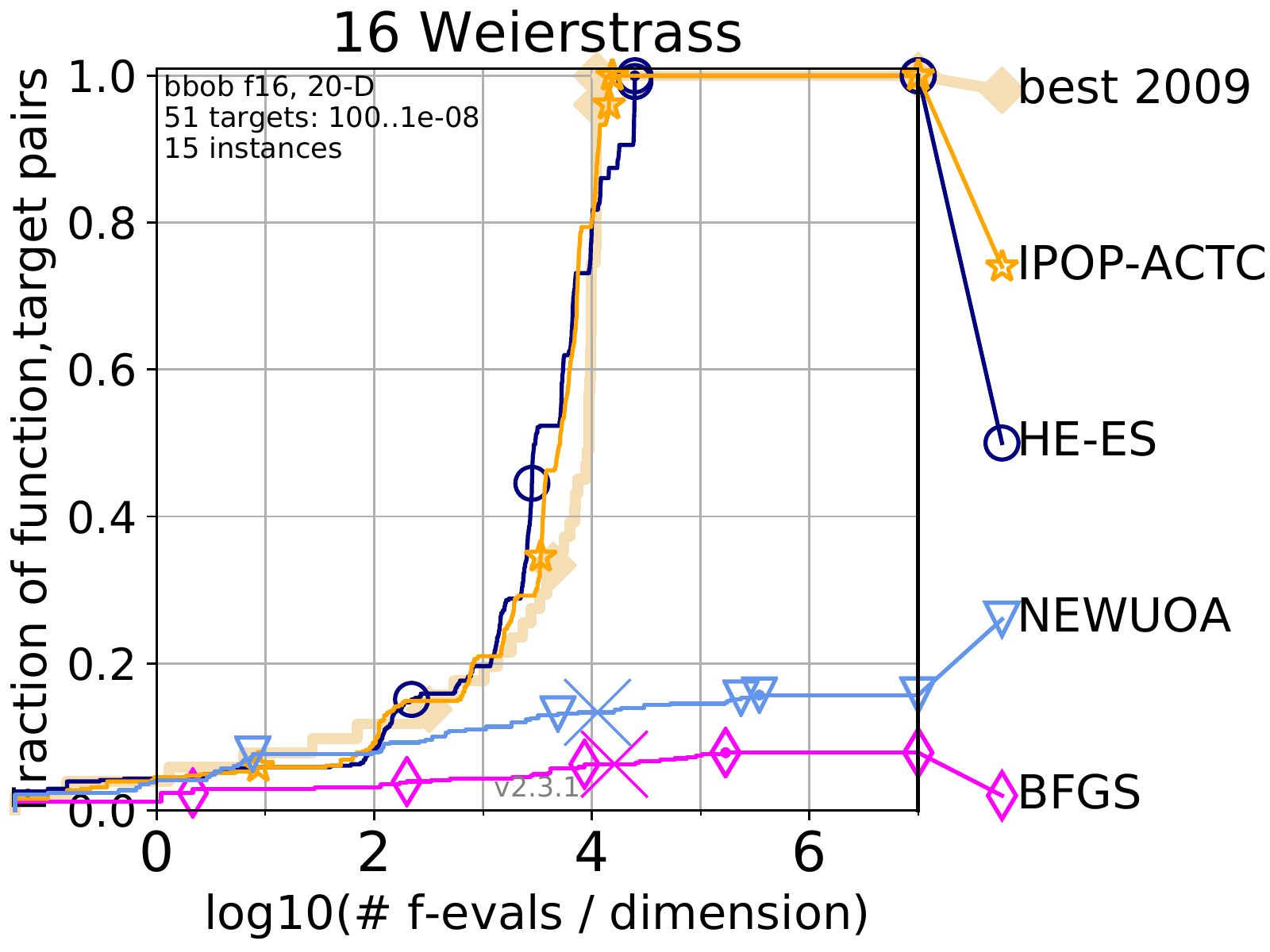}%
\includegraphics[width=0.32\textwidth]{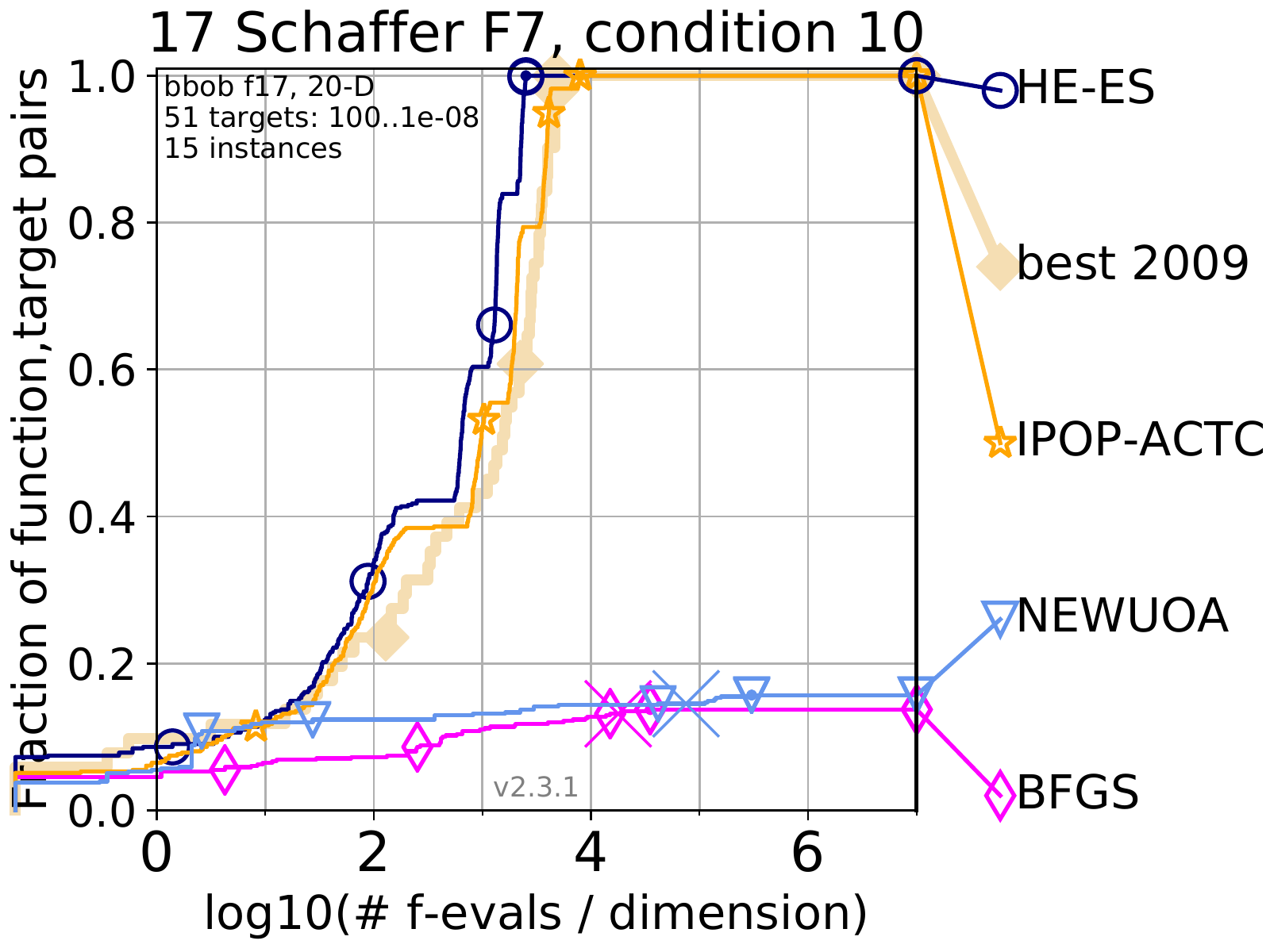}%
\includegraphics[width=0.32\textwidth]{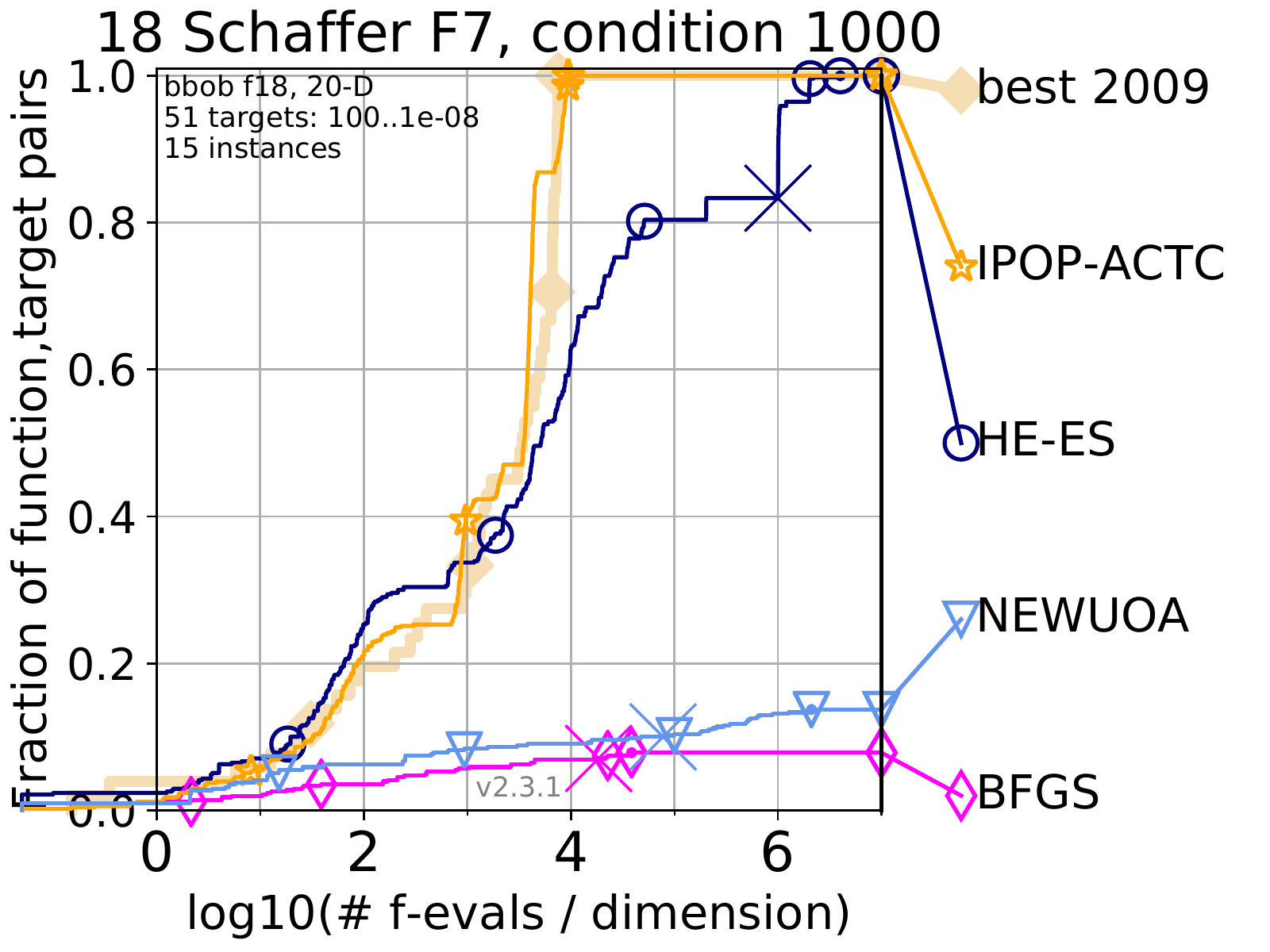}%
\\
\includegraphics[width=0.32\textwidth]{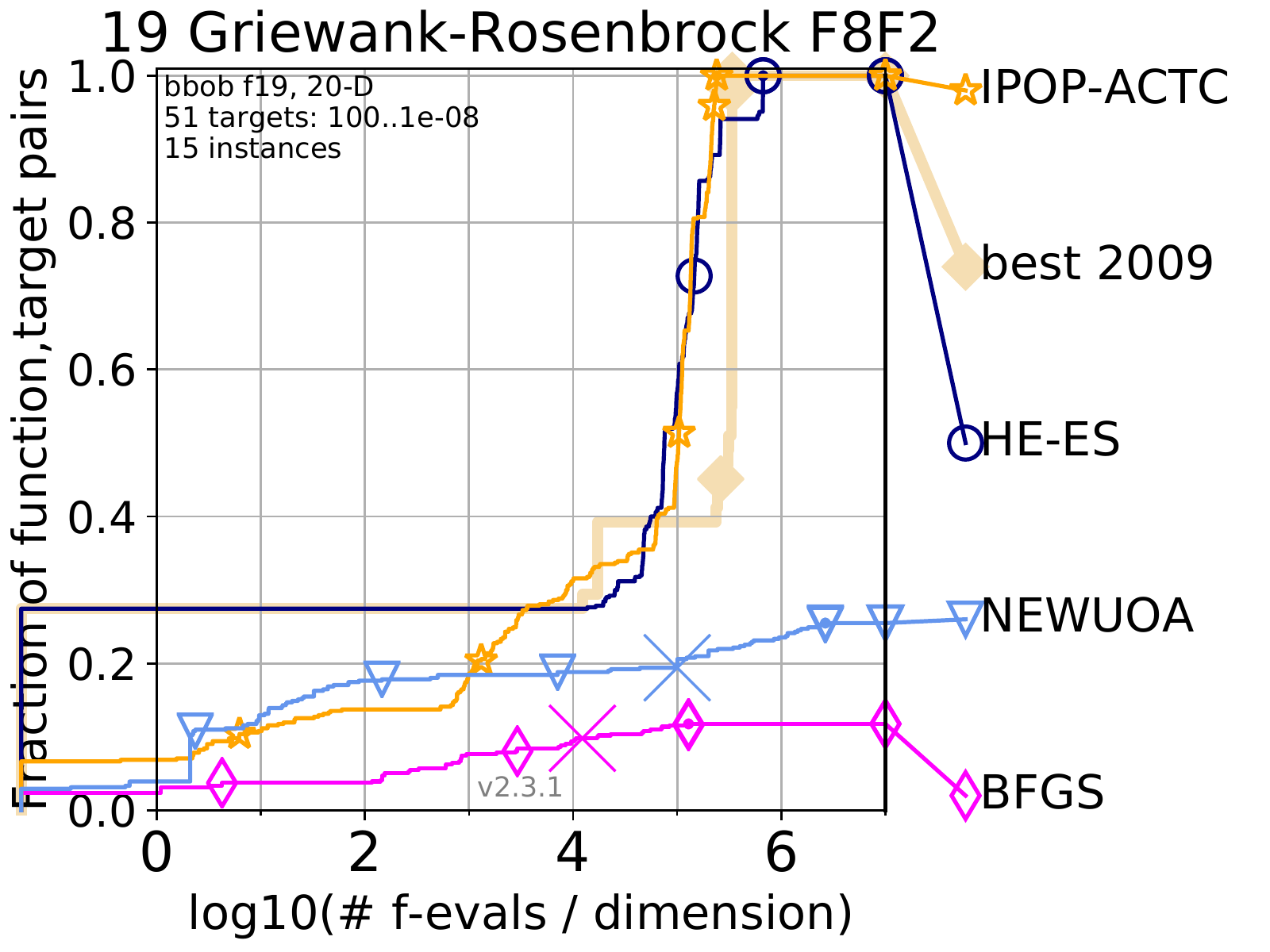}%
\includegraphics[width=0.32\textwidth]{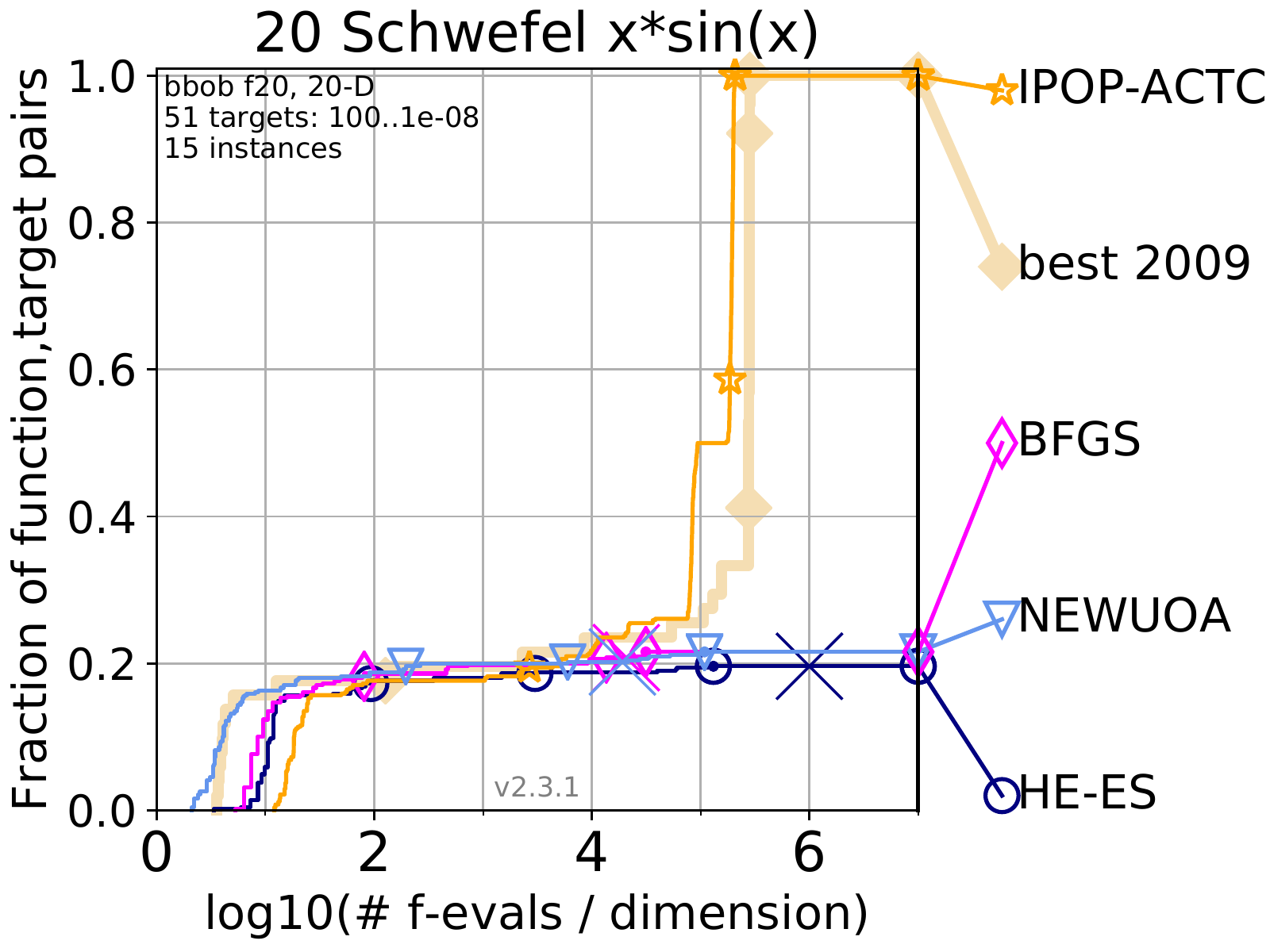}%
\includegraphics[width=0.32\textwidth]{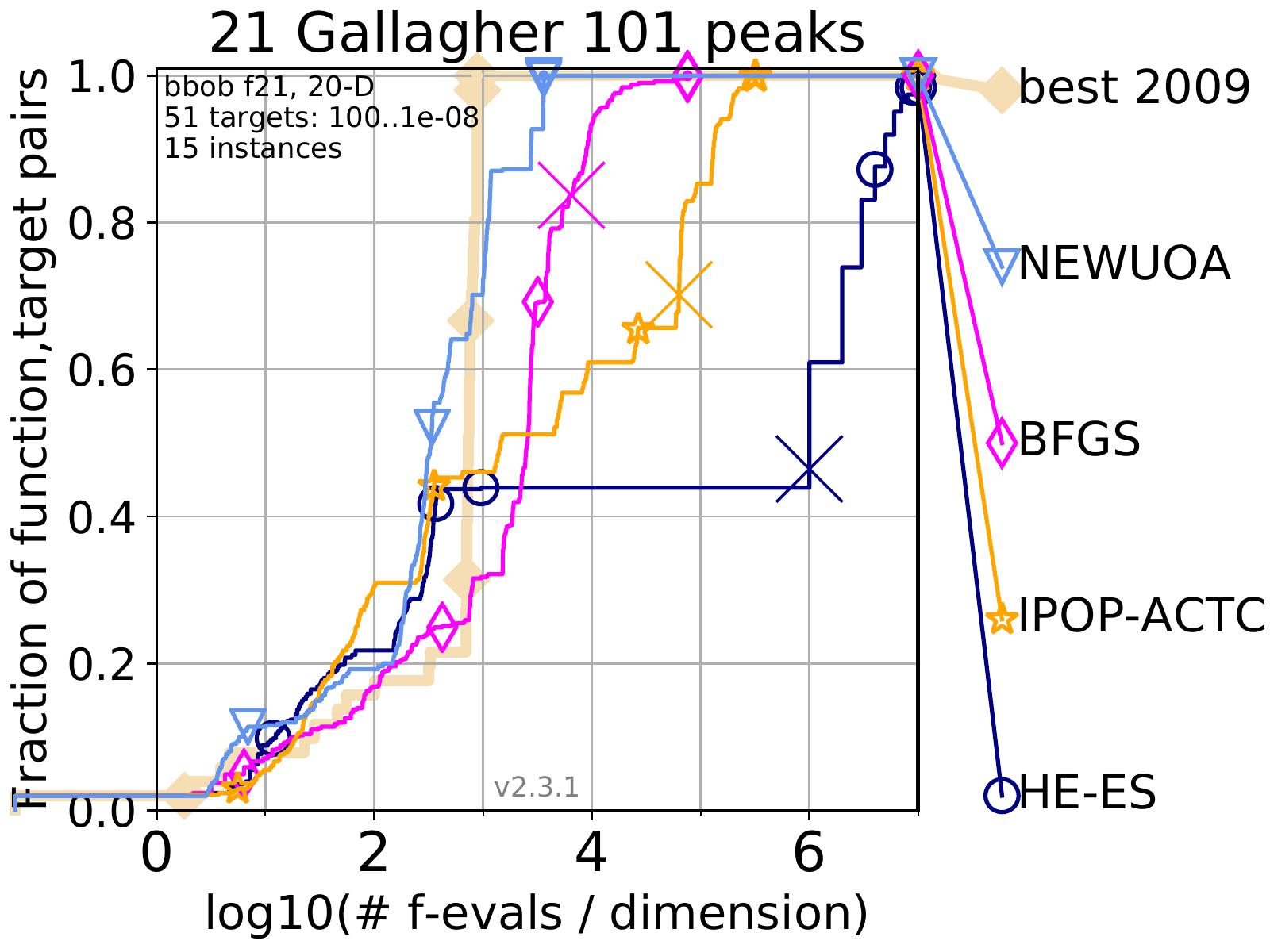}%
\\
\includegraphics[width=0.32\textwidth]{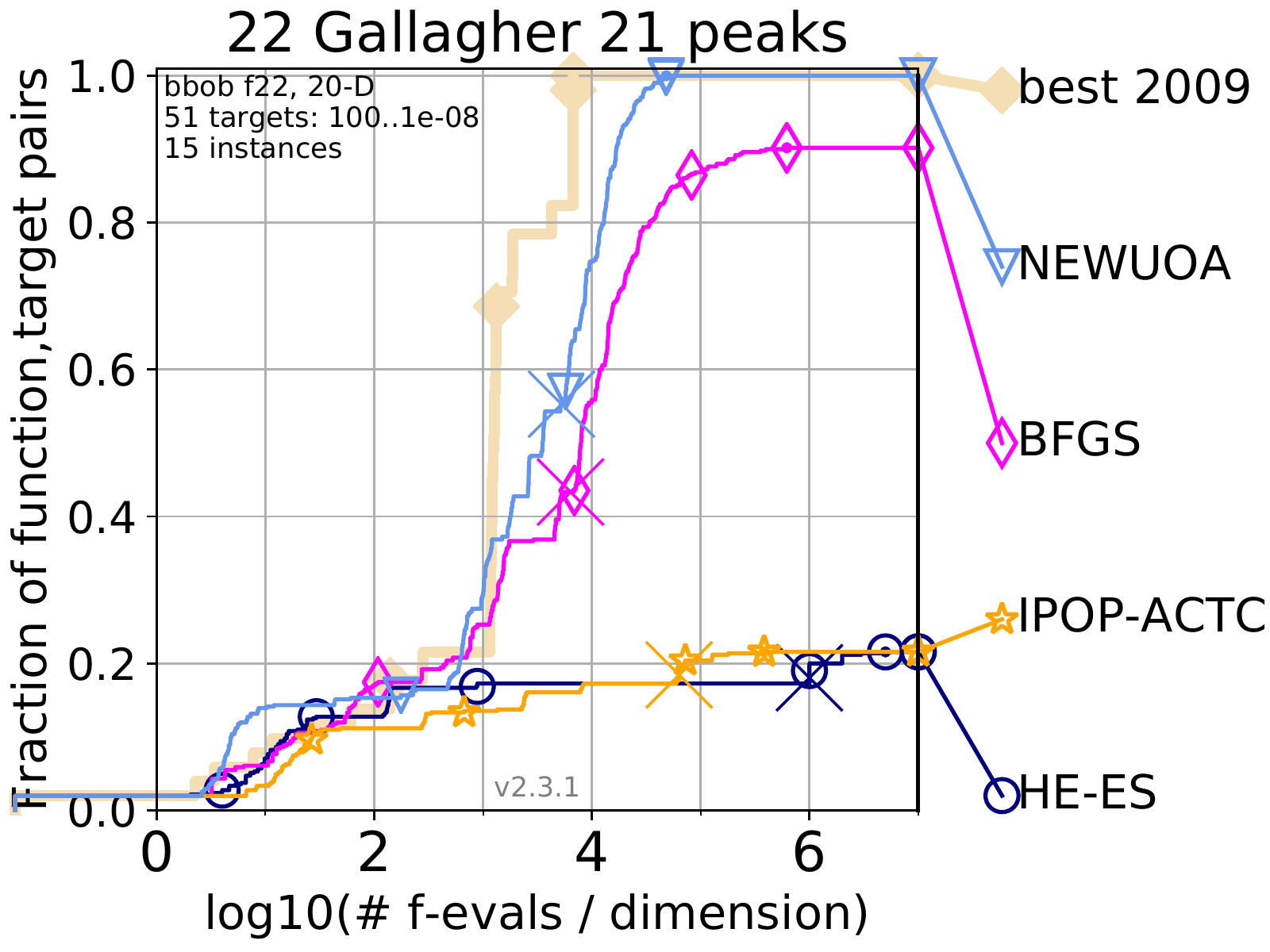}%
\includegraphics[width=0.32\textwidth]{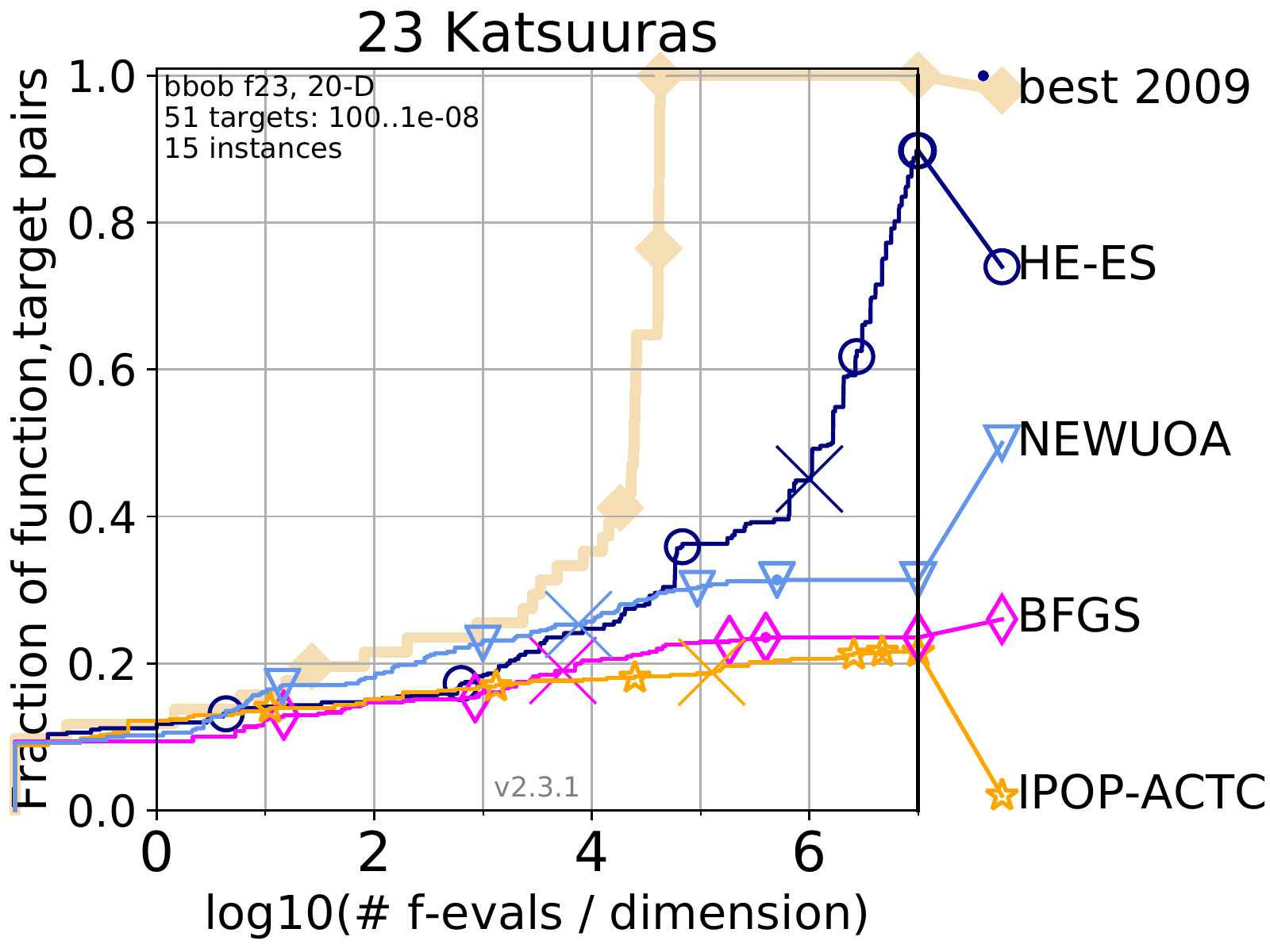}%
\includegraphics[width=0.32\textwidth]{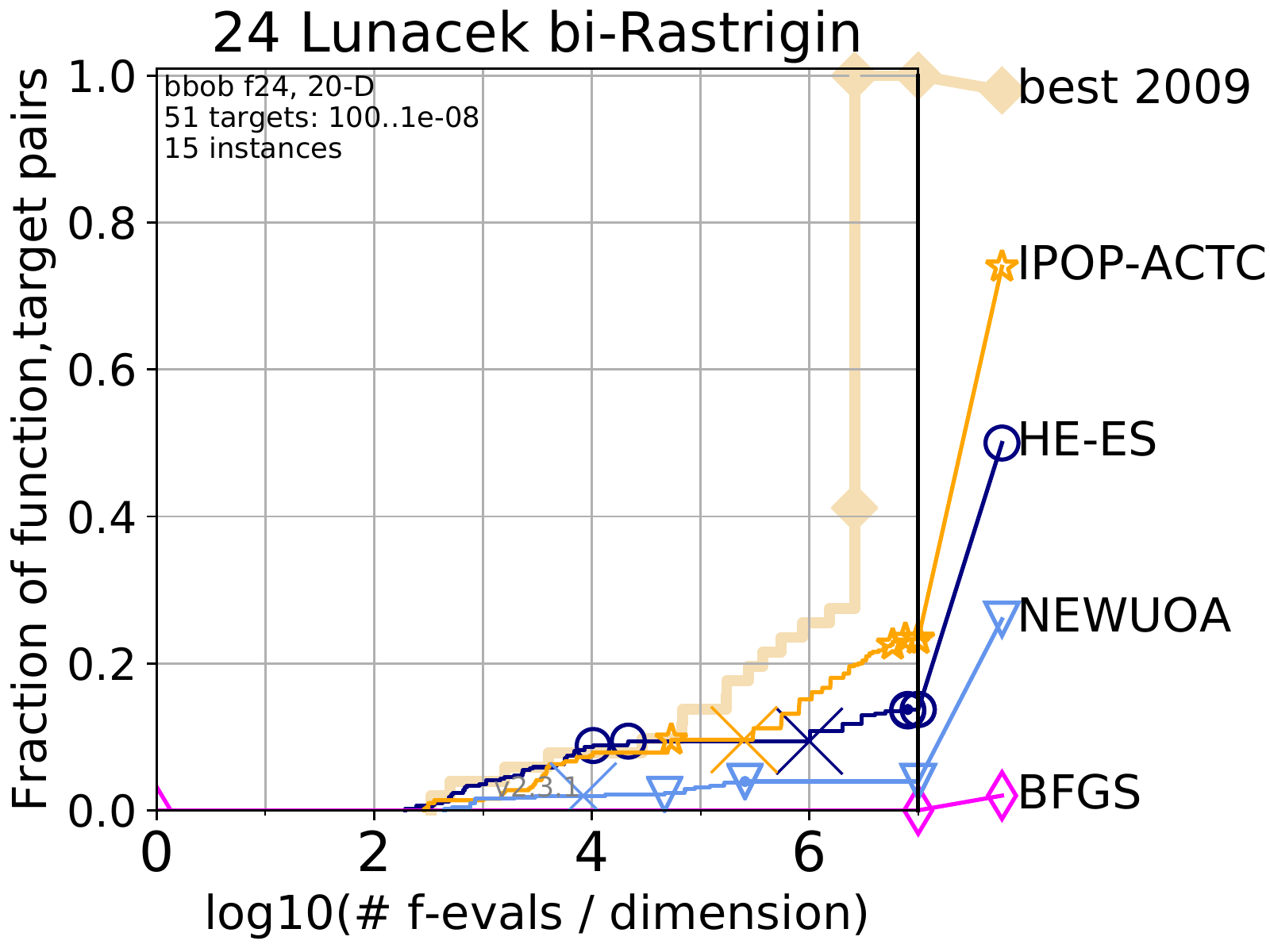}%
\caption{
	ECDF plots for the noiseless BBOB problems 16--24 in dimension $d=20$.
\label{figure:ecdf-2}}
\end{figure*}

\begin{figure*}
\includegraphics[width=0.48\textwidth]{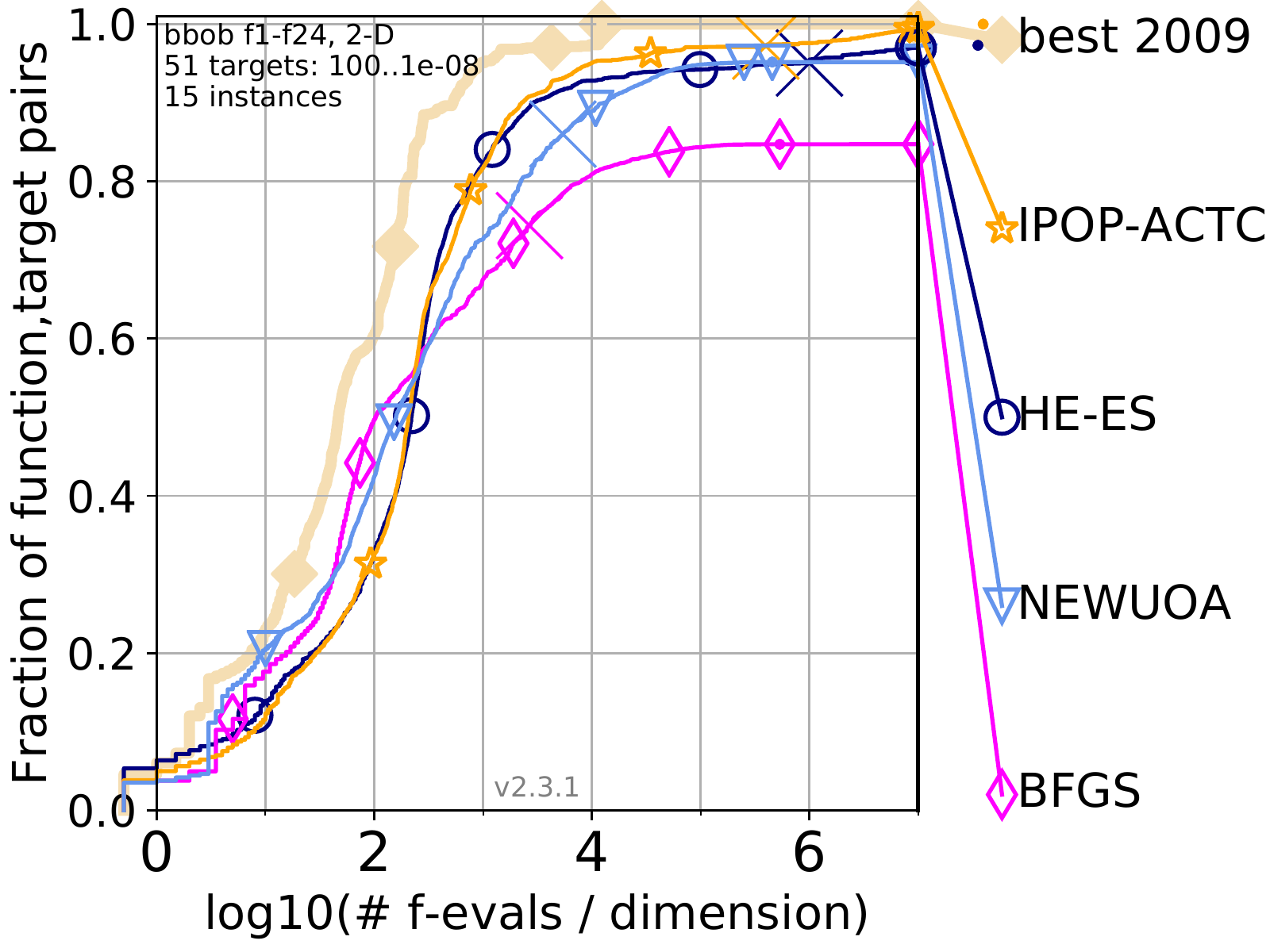}
\includegraphics[width=0.48\textwidth]{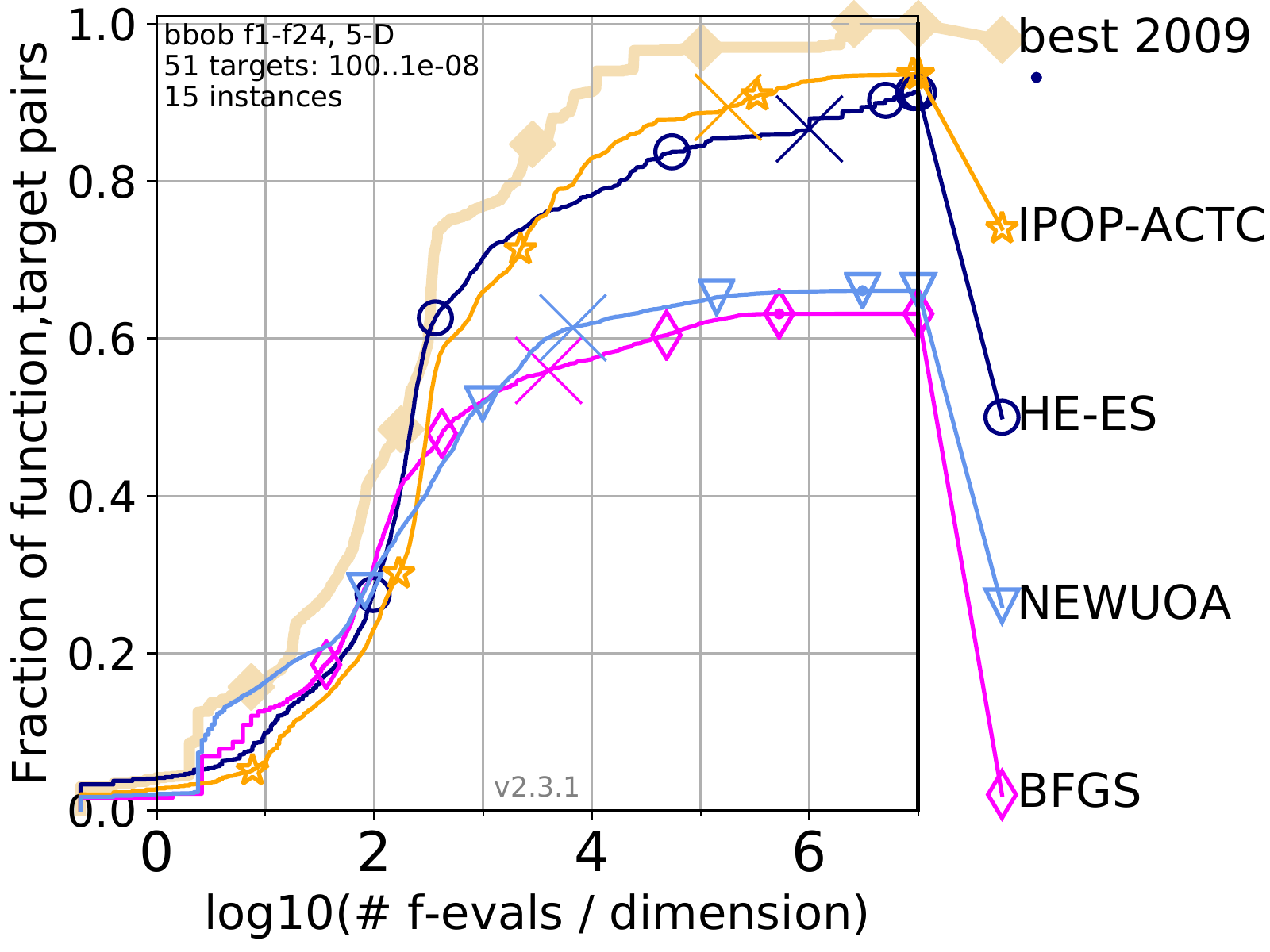}
\\
\includegraphics[width=0.48\textwidth]{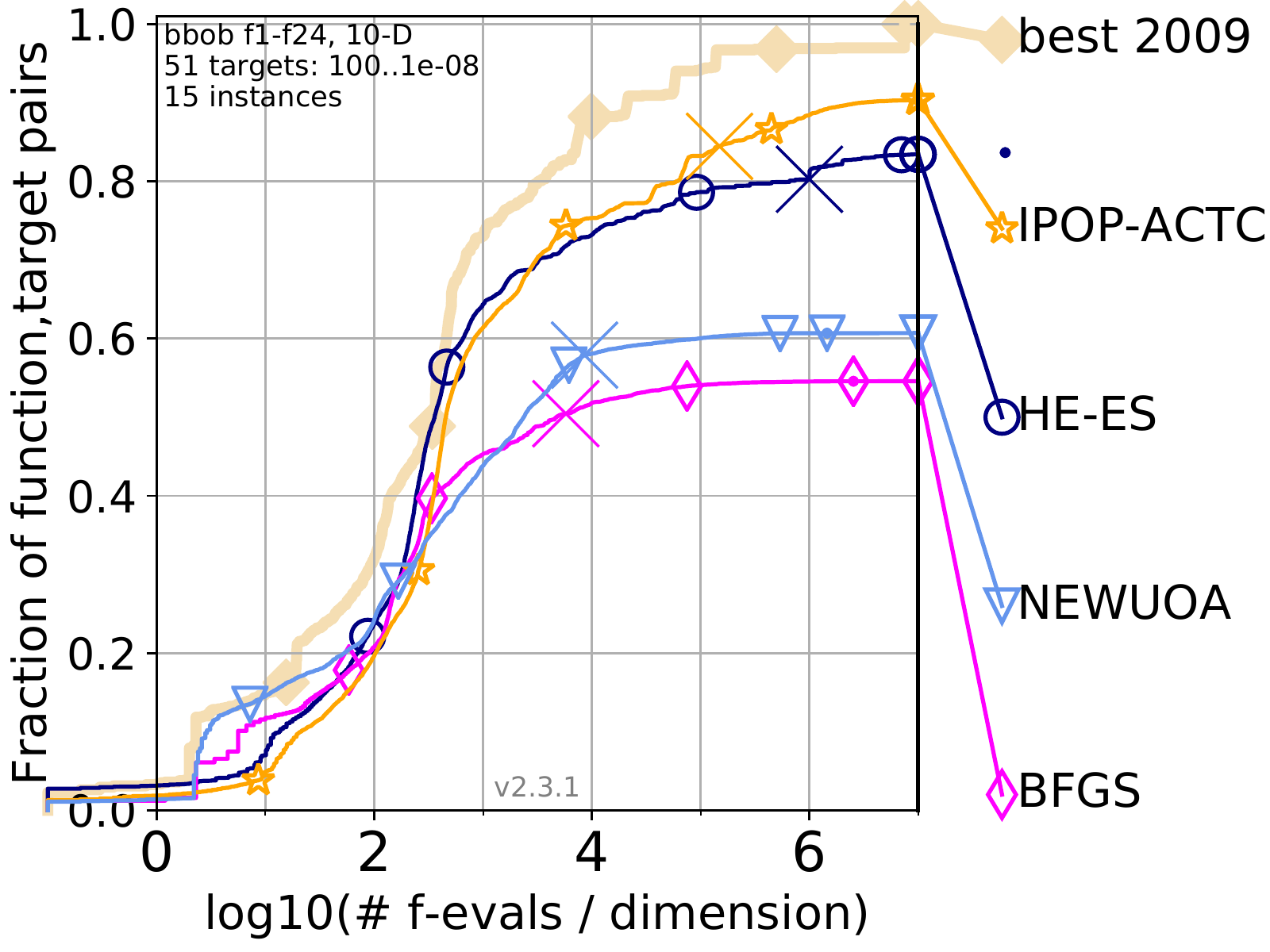}
\includegraphics[width=0.48\textwidth]{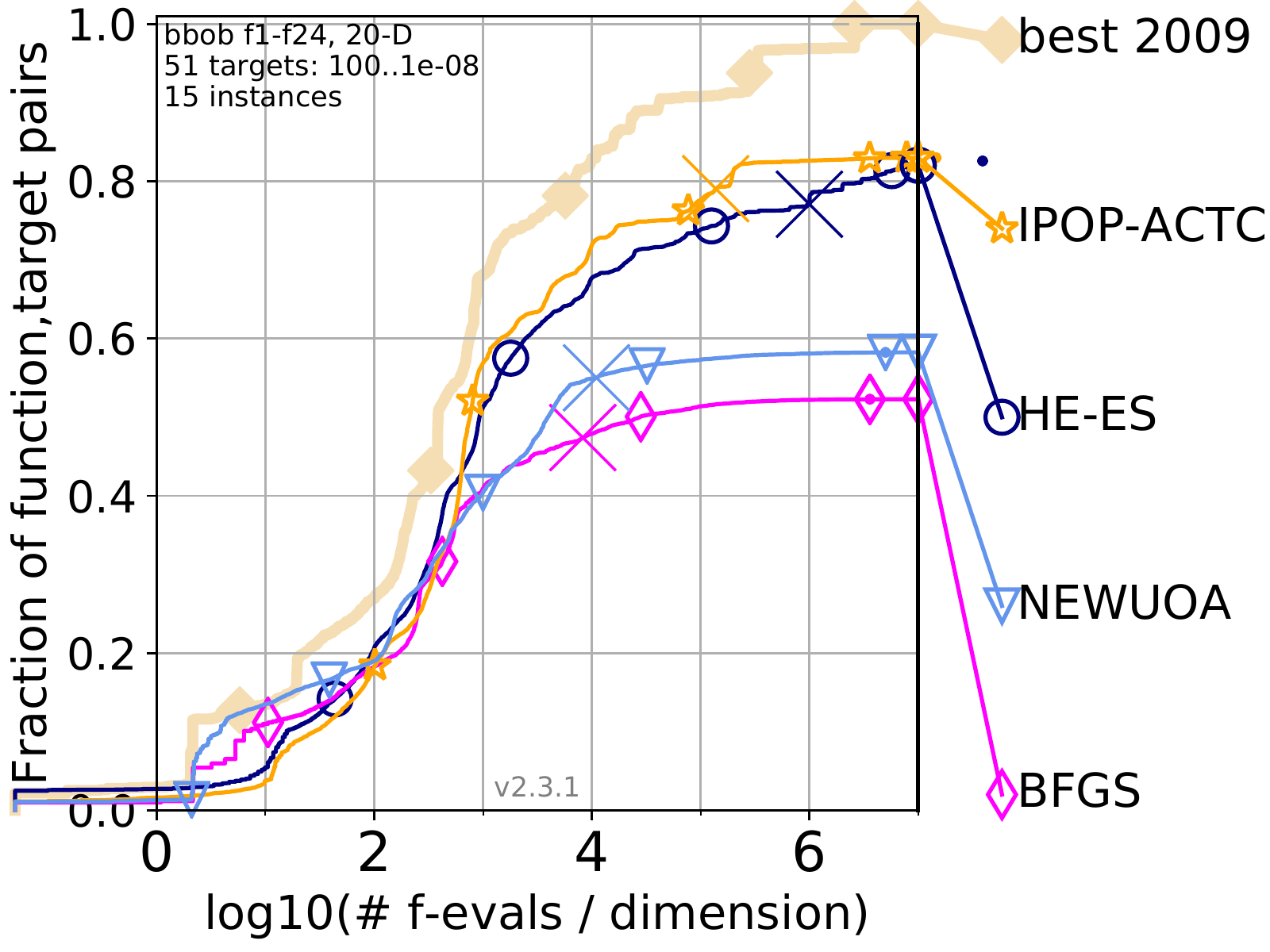}
\caption{
	Aggregated performance on all 24 BBOB functions in dimensions 2, 5,
	10 and 20. We observe that HE-ES clearly outperforms NEWUOA and
	BFGS. IPOP-CMA-ES is more reliable then HE-ES, and this gap slightly
	increases with increasing dimension. The differences mostly originate
	from hard multi-modal problems.
\label{figure:all}}
\end{figure*}

The BBOB platform generates a plethora of results. Due to space
constraints we show a representative subset thereof.
Figures \ref{figure:ecdf} and ~\ref{figure:ecdf-2} show ECDF plots on
all 24 function for problem dimension 20, with IPOP-CMA-ES, BFGS, and
NEWUOA as baselines.
Figure~\ref{figure:all} shows overall performance in dimensions 2, 5,
10, and 20. The results for IPOP-CMA-ES, BFGS, and NEWUOA were obtained
from the BBOB/COCO platform.

\begin{figure*}
\includegraphics[width=0.24\textwidth]{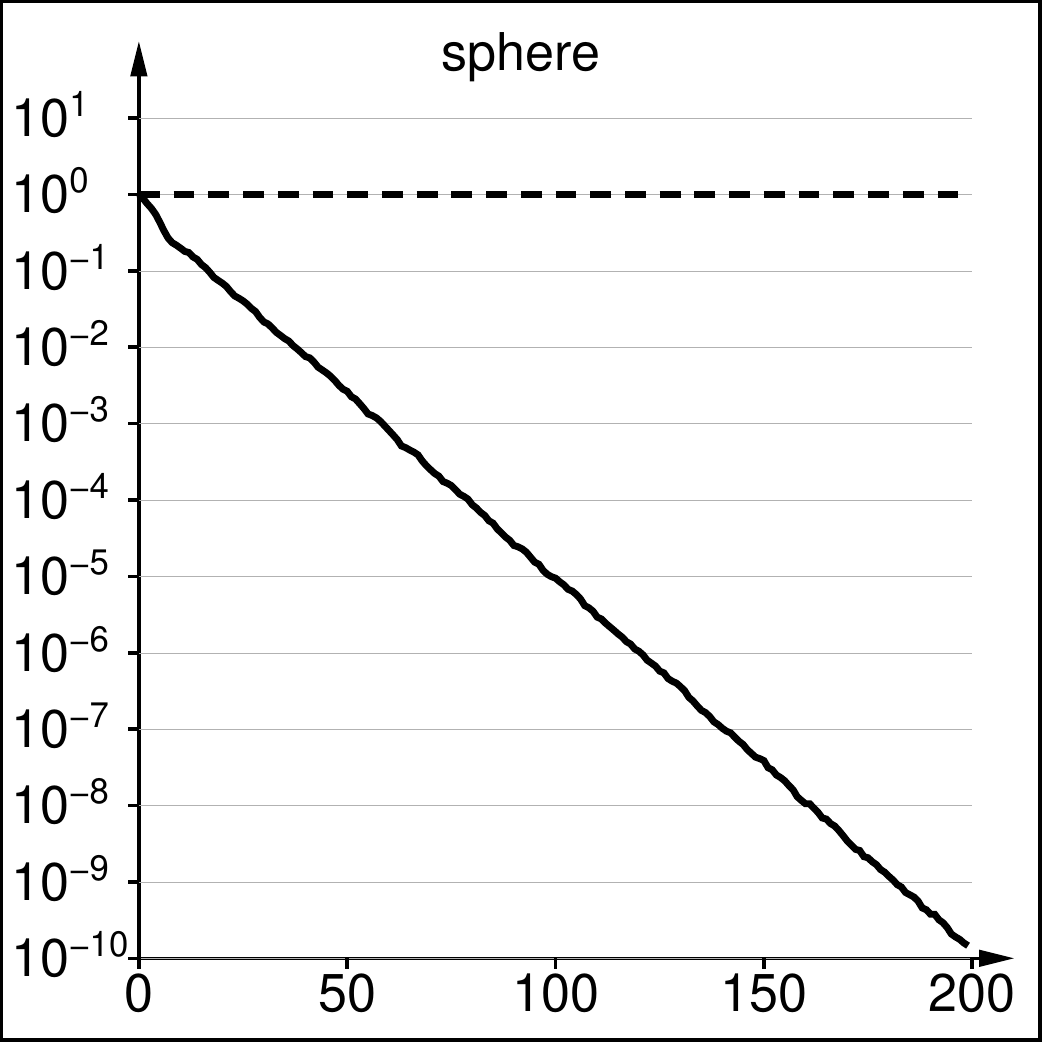}
\includegraphics[width=0.24\textwidth]{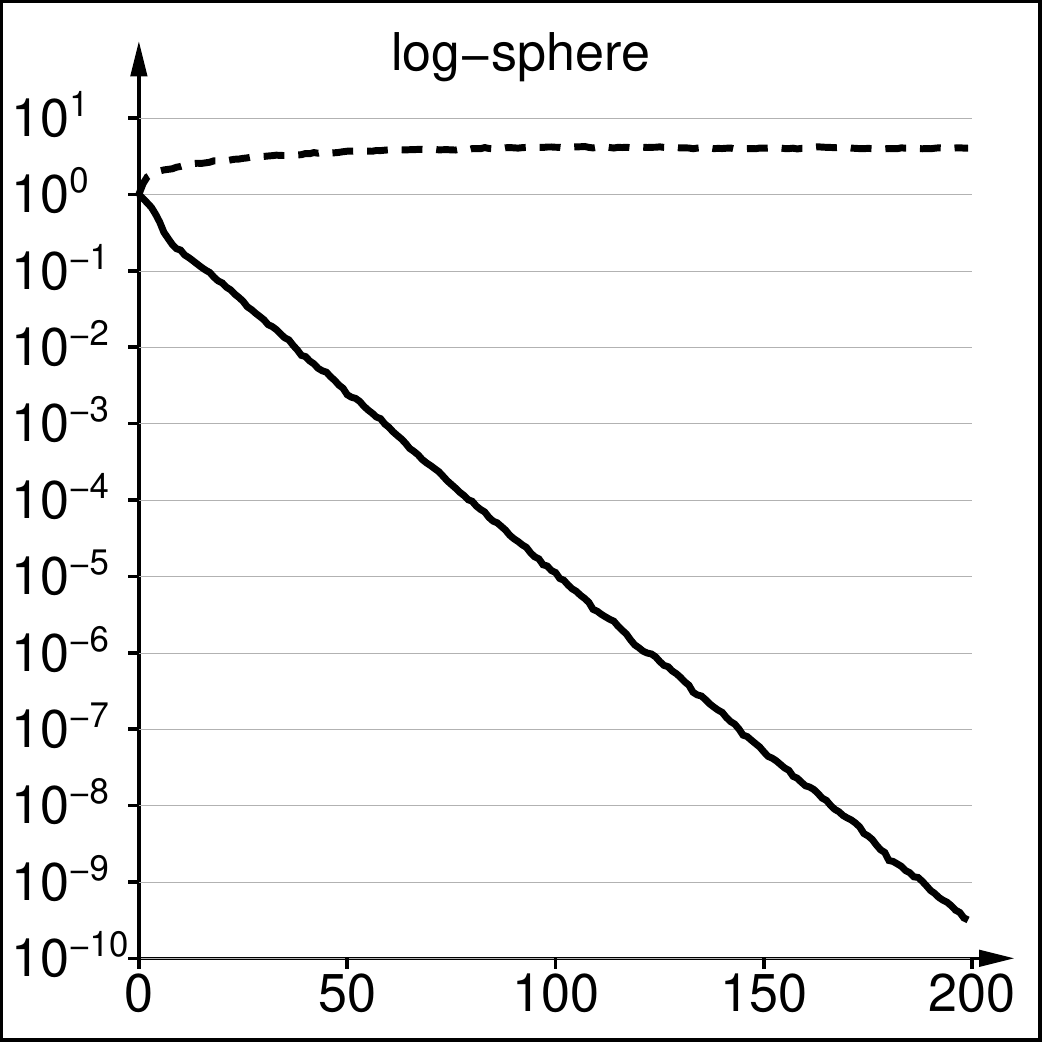}
\includegraphics[width=0.24\textwidth]{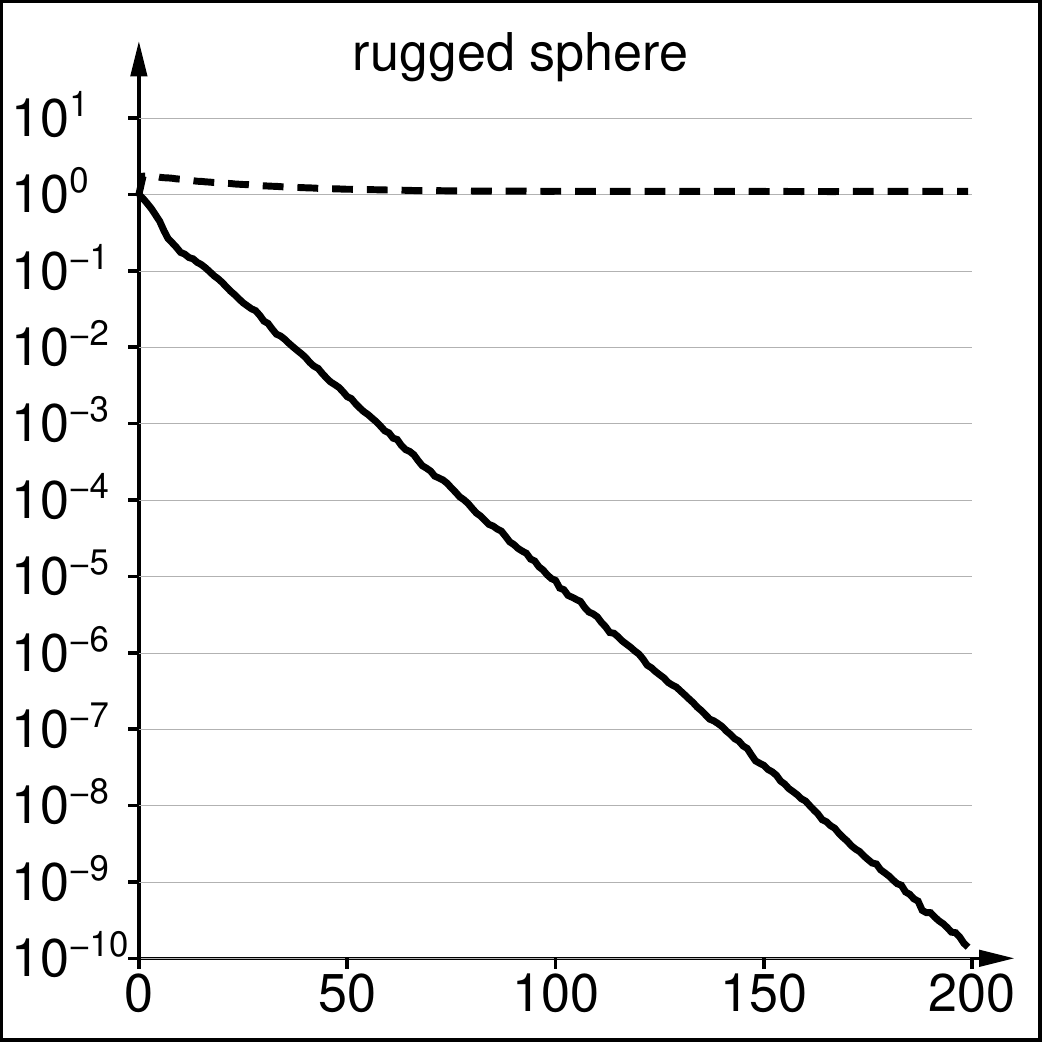}
\includegraphics[width=0.24\textwidth]{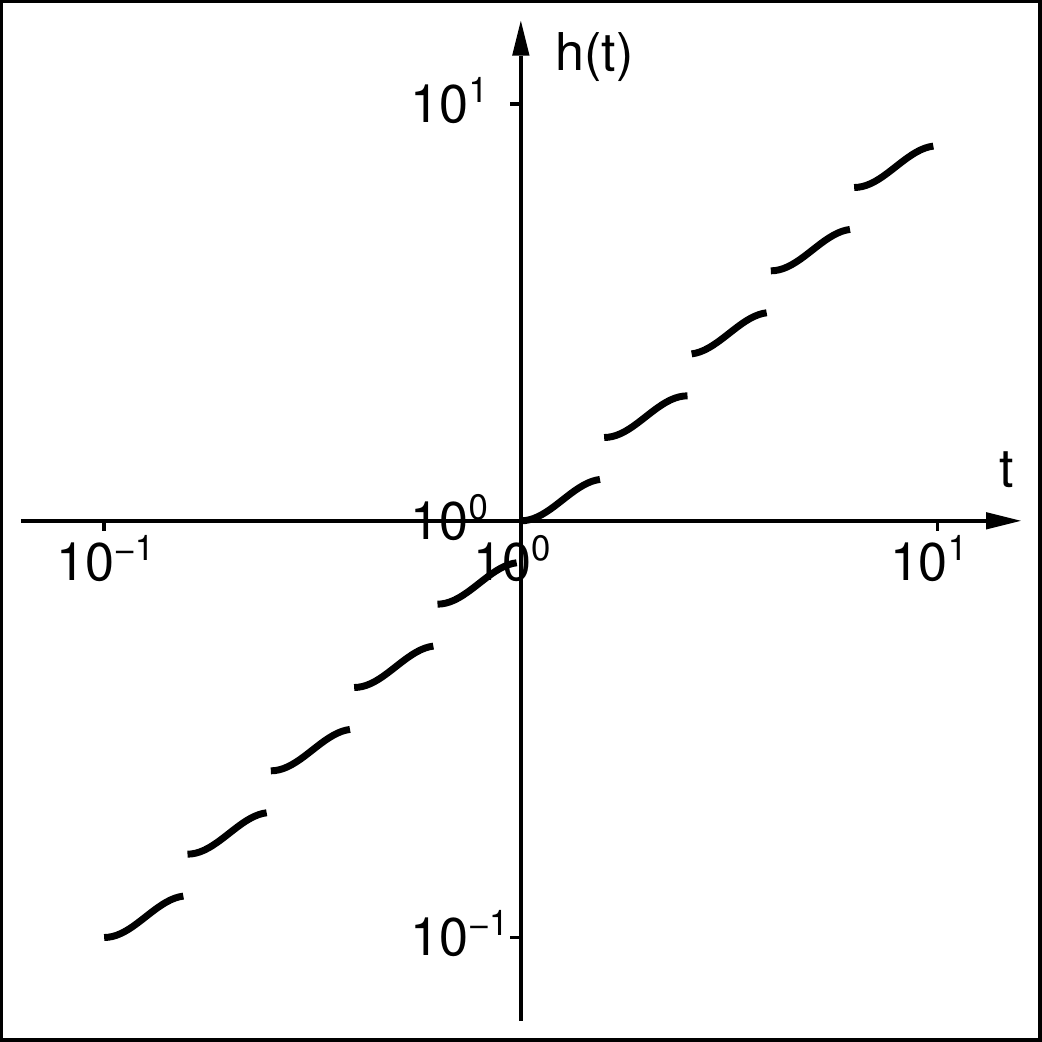}
\caption{
Distance to the optimum (solid) and condition number of $C$ (dashed)
over 200 generations of HE-ES started at $m=(1,0,\dots,0) \in \R^{10}$
with $\sigma=0.1$ for sphere, log-sphere, and rugged sphere. The curves
are medians over 99 independent runs.
Right: log-log plot of the transformation $h$. 
\label{figure:spheres}}
\end{figure*}

\paragraph{Discussion.~}

We observe excellent performance across most problems. On all convex
quadratic problems ($f_1$, $f_2$, $f_{10}$, $f_{11}$, $f_{12}$) HE-ES
performs very well, and on $f_{10}$ (ellipsoid) and $f_{11}$ (discus)
it even outperforms the hypothetical ``2009-best portfolio algorithm'',
which picks the best optimizer from the 2009 competition for each
problem. Surprisingly, the same holds for problems $f_{15}$ (Rastrigin),
$f_{16}$ (Weierstra\ss, see below), $f_{17}$ (Schaffer F7, condition 10),
and $f_{19}$ (Griewank-Rosenbrock).
Overall, the performance is much closer to CMA-ES than to NEWUOA and BFGS,
which indicates that the character of an ES is preserved, despite the
novel mechanism for updating the covariance matrix.

Compared to IPOP-CMA-ES, we observe degraded performance on $f_6$
(attractive sector), $f_{13}$ (sharp ridge), $f_{20}$ (Schwefel
$x\cdot\sin(x)$), $f_{21}$ and $f_{22}$ (Gallagher peaks), $f_{23}$
(Katsuuras), and $f_{24}$ (Lunacek bi-Rastrigin). HE-ES apparently
struggles with the asymmetry of the attractive sector problem, which
can yield drastically wrong curvature estimates. Similarly, it is
conceivable that estimating curvatures on the sharp ridge problem is
prone to failure. We believe that these two benchmark functions
highlight inherent limitations of the HE-ES update.

\paragraph{Control Experiments.~}

In order to understand the weak performance on the highly multi-modal
problems despite IPOP restarts we investigated the behavior of HE-ES on the
deceptive Lunacek problem $f_{24}$ in dimension $d=10$. We ran HE-ES with
IPOP restarts 100 times with a reduced budget of $10^4 \cdot d$ function
evaluations. This medium-sized budget is $100$ times smaller than in the
BBOB/COCO experiments. It suffices for 4 to 5 runs, with population sizes
ranging from $10$ to $160$. We found that HE-ES converged to the better of
the two funnels in 83 cases, and solved the problem to a high precision of
$10^{-10}$ in 40 out of 100 cases, which corresponds to reaching all BBOB
targets. This means that the correct funnel and the best local optimum of
the Rastrigin structure were found in $40\%$ of the cases, which is a quite
satisfactory behavior. It is noteworthy that CMA mechanisms are not even
needed for this problem (indeed, performance is unchanged when disabling
CMA), and hence the performance difference to CMA-ES is probably an
artifact of a different restart implementation. This is unrelated to the
new CMA mechanism and hence of minor relevance for our investigation.

As mentioned above, HE-ES performs surprisingly well on the Weierstra{\ss}
function $f_{16}$, which is continuous but nowhere differentiable, and
therefore strongly violates the assumption of a twice continuously
differentiable objective function. At first glance, this result is
surprising. The reason is that HE-ES does not really need correct
estimates of the curvature. It is only relevant that the structure
(the ``global trend'') of the objective function at the relevant scale
(given by $\sigma$) is correctly captured. Of course, HE-ES picks up
misleading curvature information. However, since the Weierstra{\ss} function
does not exhibit a systematic preference for a particular direction,
such unhelpful information averages out over time and hence does not
have a lasting detrimental effect.

In order to investigate this effect closer we performed the following
experiment. We start from the 10-dimensional sphere function
$f(x) = \frac12 \|x\|^2$ as a base case. Then we create two variants by
monotonically transforming the function values, leaving the level sets
intact, resulting in the non-convex function $\log(f(x))$ (log-sphere),
and the rugged and discontinuous function $h(f(x))$ with
$h(t) = \exp\left (\left[ \frac14 - \frac12 \cos(\pi (5 \log(t)-r(t))) + r(t) \right] / 5 \right)$,
where $r(t) = \lfloor 5 \log(t) \rfloor$ (rugged sphere).
Figure~\ref{figure:spheres} shows a plot of the transformation $h$ as
well as the resulting optimization performance of HE-ES. For sphere the
condition number remains at exactly one (the optimal value).
Importantly, in the other cases the condition number remains close
enough to one so as to not impair optimization performance. For the
log-sphere there is a slowdown, however, of negligible magnitude: HE-ES
requires about $5\%$ more time. We observe that in this setup, surprisingly,
HE-ES suffers nearly not at all from misleading curvature estimates.

\section{Conclusion}

We have presented the Hessian Estimation Evolution Strategy (HE-ES), an
ES with a novel covariance matrix adaptation mechanism.
It adapts the covariance matrix towards the inverse Hessian projected to
random lines, estimated through finite differences by means of mirrored
sampling. The algorithm comes with a specialized cumulative step size
adaptation rule for mirrored sampling.

Despite its seemingly strong assumptions the method works well on a broad
range of problems. It is particularly well suited for smooth unimodal
problems like convex quadratic functions and the Rosenbrock function.
Surprisingly, the adaptation mechanism that is based on estimating
presumedly positive second derivatives can work well on non-convex
and even on discontinuous problems.

We believe that the HE-ES offers an interesting alternative to adapting
the covariance matrix of the sampling distribution towards the maximum
likelihood estimator of successful steps, corresponding to the natural
gradient in parameter space.

\end{document}